\documentclass[iicol]{sn-jnl} 

\usepackage{graphicx}%
\usepackage{multirow}%
\usepackage{amsmath,amssymb,amsfonts}%
\usepackage{amsthm}%
\usepackage{mathrsfs}%
\usepackage[title]{appendix}%
\usepackage{xcolor}%
\usepackage{textcomp}%
\usepackage{manyfoot}%
\usepackage{booktabs}%
\usepackage{algorithm}%
\usepackage{algorithmicx}%
\usepackage{algpseudocode}%
\usepackage{listings}
\usepackage{subcaption}
\usepackage{apacite}
\usepackage{colortbl}

\newcommand{\tb}[1]{\textcolor{black}{#1}}

\newcommand{\h}[1]{\mathbf{#1}}
\makeatletter
\renewcommand{\maketag@@@}[1]{\hbox{\m@th\normalsize\normalfont#1}}%
\makeatother

\newtheorem{theorem}{Theorem}
\newtheorem{proposition}[theorem]{Proposition}%
\newtheorem{remark}{Remark}%

\newtheorem{definition}{Definition}%

\begin{document}

\title[Article Title]{
Hyperspectral and multispectral image fusion with arbitrary resolution through self-supervised representations}



\author[1,]{\fnm{Ting} \sur{Wang}}\equalcont{Equal contributions.}

\author[2,]{\fnm{Zipei} \sur{Yan}}
\equalcont{Equal contributions.}

\author[2]{\fnm{Jizhou} \sur{Li}}

\author[3]{\fnm{Xile} \sur{Zhao}}

\author*[1]{\fnm{Chao} \sur{Wang}}\email{wangc6@sustech.edu.cn}

\author[4]{\fnm{Michael} \sur{Ng}} 




\affil*[1]{Department of Statistics and Data Science, Southern University of Science and Technology, Shenzhen, P.R. China}
 \affil[2]{Department of Electronic Engineering, The Chinese University of Hong Kong, Hong Kong SAR, P.R. China}
\affil[3]{School of Mathematical Science, University of Electronic Science and Technology of China, Chengdu, P.R. China}
\affil[4]{Department of Mathematics, Hong Kong Baptist University, Hong Kong SAR, P.R. China}

\abstract{
The fusion of a low-resolution hyperspectral image (LR-HSI) with a high-resolution multispectral image (HR-MSI) has emerged as an effective technique for achieving HSI super-resolution (SR). Previous studies have mainly concentrated on estimating the posterior distribution of the latent high-resolution hyperspectral image (HR-HSI), leveraging an appropriate image prior and likelihood computed from the discrepancy between the latent HSI and observed images. Low rankness stands out for preserving latent HSI characteristics through matrix factorization among the various priors. { 
However, the primary limitation in previous studies lies in the generalization of a fusion model with fixed resolution scales, which necessitates retraining whenever output resolutions are changed. }To overcome this limitation, we propose a novel continuous low-rank factorization (CLoRF) by integrating two neural representations into the matrix factorization, capturing spatial and spectral information, respectively. This approach enables us to harness both the low rankness from the matrix factorization and the continuity from neural representation in a self-supervised manner. Theoretically, we prove the low-rank property and Lipschitz continuity in the proposed continuous low-rank factorization. Experimentally, our method significantly surpasses existing techniques and achieves user-desired resolutions without the need for neural network retraining.  {Code is available at \href{https://github.com/wangting1907/CLoRF-Fusion}{https://github.com/wangting1907/CLoRF-Fusion}. }
}




\maketitle

\section{Introduction}

\begin{figure*}[t!]
    \centering
    \includegraphics[width=1.02\textwidth]{ 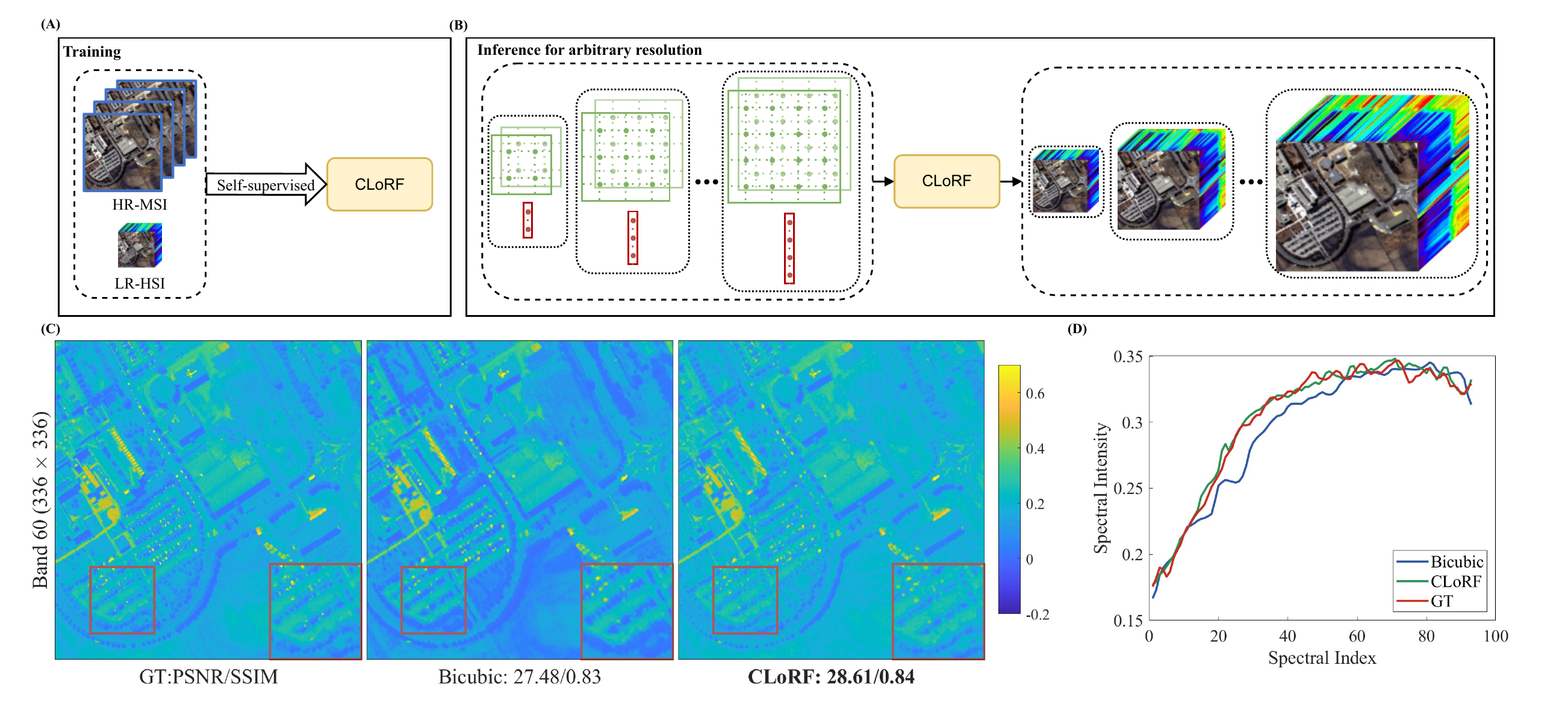} 
    \caption{The pipeline of CLoRF for arbitrary resolution. (A) Train the CLoRF. (B) Use the trained CLoRF to infer arbitrary resolutions of HSIs with given spatial and spectral coordinates. 
    (C) 
    An example of CLoRF for super-resolution on the Pavia University ($336\times336\times93$). The CLoRF is trained given its LR-MSI ($168\times168\times4$) and HR-HSI ($42\times42\times50$), then infers the original resolution. Bicubic interpolation directly upsamples from LR. \tb {In the spectral domain, each band has distinct brightness values, and bicubic interpolation estimates the missing bands by referencing adjacent ones. As a result, this can cause discrepancies in the color representation of the interpolated image, such as in band 60, when compared to the GT image.} (D) Visualize the spectrum of a random pixel from the results on (C).}
    \label{fig:aribrtary}
\end{figure*}

Hyperspectral images (HSIs) have widespread applications across various fields due to their rich spectral information. The abundant spectral details offered by HSIs facilitate accurate scene interpretation and enhance the efficacy of numerous applications, including object classification~\cite{gao2014subspace} and anomaly detection~\cite{guo2014weighted}. However, the inherent trade-off between spectral and spatial resolution in HSI systems, constrained by hardware limitations, often results in HSIs with lower spatial resolution than RGB, panchromatic (PAN), and multispectral images (MSI). To enhance the spatial resolution of HSIs, a natural approach is to fuse LR-HSI and HR-MSI, known as hyperspectral and multispectral image fusion (HSI-MSI fusion). HSI-MSI fusion resembles MSI pansharpening, where low spatial resolution MSI is merged with high-resolution PAN imagery. However, directly applying these pansharpening methods to fuse HSI and MSI images suffers from challenges, as PAN images have limited spectral information, leading to spectral distortion~\cite{loncan2015hyperspectral}. Consequently, numerous approaches tailored for HSI-MSI fusion are introduced, which can be generally categorized into model-based methods and deep learning-based models.

\begin{figure*}[t!]
\begin{center}
\includegraphics[width=1\textwidth]{ 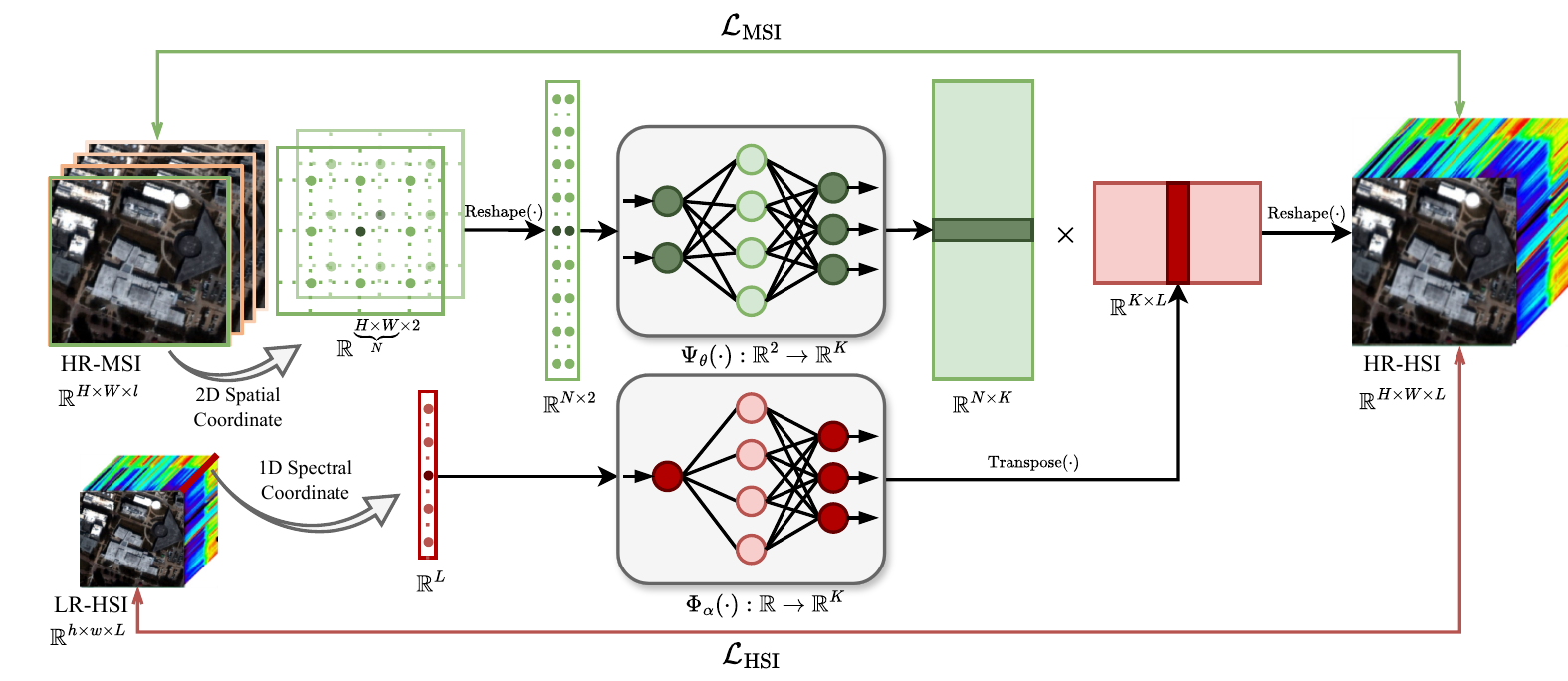} 
\end{center}
\caption{Illustration of the proposed CLoRF for MSI-HSI fusion. The spatial coordinates and spectral coordinates of HR-MSI and LR-HSI are fed into the Spatial-INR $\Phi_{\theta}(\cdot)$ and Spectral-INR $\Psi_{\alpha}(\cdot)$ to generate coefficients and the bases, respectively. Thereafter, the generated coefficients and the bases are multiplied to recover the HR-HSI. 
}
\label{flowchart}
\end{figure*}

Model-based methods leverage the low-rank structure of HSIs by characterizing their low-rankness through matrix factorization, decomposing the HSI matrix into the basis and coefficients or endmembers and abundances. Therefore, the principles of matrix factorization-based methods rely on appropriate prior information and the likelihood determined by the relationships between the latent HSI and the observed LR-HSI and HR-MSI. Given the known degradation model, most existing methods focus on modeling prior information for HSIs, including explicit and implicit methods. The explicit methods employ handcrafted explicit prior information, such as low-rankness~\cite{zhang2018exploiting,wang2020hyperspectral}, smoothness~\cite{simoes2014convex}, sparsity~\cite{dong2016hyperspectral}, and non-local similarity~\cite{dian2017hyperspectral}, to depict the prior distribution of the latent HSI for super-resolution (SR). Implicit smooth regularizations introduce basis functions to parameterize the prior information, extending the modeling to functional representations. For example, \citeauthor{yokota2015smooth,debals2017nonnegative} utilize non-negative matrix factorization parameterized by basis functions to reveal implicit smoothness.

{Compared to matrix-based methods, tensor-based methods can directly process HSI data and have garnered significant attention~\cite{dian2019nonlocal, 9556548}. For example, a nonlocal sparse tensor factorization method based on Tucker decomposition was introduced in~\cite{9328229}, which factorizes an HSI into a sparse core tensor multiplied by dictionaries along both the spatial and spectral dimensions. In~\cite{10560507}, a novel fusion model is proposed within the tensor ring decomposition framework, rather than using subspace decomposition-based fidelity terms. Although these tensor decomposition-based methods are effective in preserving the spectral-spatial correlation of HSI, they face the challenge of dimensionality catastrophe.}
Furthermore, with the success of deep learning, \citeauthor{dian2020regularizing,wang2023self} implicitly introduce a denoising operator into the optimization framework to learn the deep image prior, which is shared across all HSIs using deep neural networks with different structures. Although these methods have achieved significant success in HSI-MSI fusion, manually designed explicit prior information may make the regularization terms problematic, resulting in significant computational complexity. Meanwhile, implicitly designed prior information sometimes becomes inappropriate for capturing the complex and detailed structures of HSIs. Moreover, these methods are restricted to fixed resolution and cannot be instantly applied to recovering HSIs with arbitrary resolution.

Most deep learning-based methods typically involve supervised training on large datasets consisting of observed and ground truth data to learn a complex function mapping. Some representative deep learning methods are based on convolution neural network (CNN) for HSI-MSI fusion, such as the two-stream fusion network designed for HR-MSI and LR-HSI~\cite{wang2023mct, khader2023model, jia2023multiscale, wang2024general}. With the remarkable performance of implicit neural representations (INRs) in continuous multi-dimensional data representation, some prior research~\cite{wang2023ss,he2024two} combined INR and CNN for HSI-MSI fusion. Although these methods have demonstrated promising results in fusion, they heavily rely on high-quality training data for supervised learning. Collecting high-quality ground-truth data is extremely time-consuming and costly. In addition, unsupervised fusion methods~\cite{zheng2020coupled, nguyen2022deep, wu2024unsupervised} are introduced, eliminating the need for expensive training data. However, these methods typically rely on complex network structures. Besides, they are also confined to fixed resolution, and inferring results at the user-desired resolutions requires retraining.  {Arbitrary-resolution HSI-MSI fusion is an emerging area within the field of image fusion. Unlike traditional fusion tasks, which typically merge images at fixed resolutions, arbitrary-resolution fusion leverages both HR-MSI and LR-HSI to generate fused outputs at any desired spatial and spectral resolutions~\cite{wang2024general,he2021cnn}. This advanced fusion technique provides greater flexibility in adjusting both the spatial and spectral resolutions of the resulting HR-HSI, offering significant potential for a wide range of applications~\cite{he2021cnn}, such as object detection~\cite{qu2017does} and land use/cover classification~\cite{gao2014subspace}. However, arbitrary-resolution hyperspectral image fusion presents greater challenges compared to standard fusion methods.}


To address the aforementioned issues, we propose an unsupervised HSI-MSI fusion framework for continuous low-rank factorization (CLoRF). Specifically, CLoRF integrates two implicit neural representations into the low-rank factorization, capturing continuous spatial and spectral information of HR-MSI and LR-HSI, respectively. Unlike classic discrete matrix factorization, we define the continuous matrix function factorization following tensor function~\cite{luo2023low}, which characterizes low-rankness in continuous representation. Each continuous function is a realization of INR parameterized by multi-layer perceptrons (MLPs). As MLPs are Lipschitz smooth, further characterizing smoothness in continuous representation. Compared to previous fusion methods, as shown in Fig.~\ref{fig:aribrtary}, CLoRF can achieve arbitrary resolution in both spatial and spectral domains without additional image information and retraining. 

Our contributions are summarized as follows:

\begin{enumerate}
    \item We propose a novel unsupervised HSI-MSI fusion method: CLoRF, which represents HSIs in a continuous representation using low-rank function factorization. CLoRF can infer with arbitrary resolution without the need for retraining.

    \item We theoretically prove that the implicit regularization terms of low-rankness and smoothness are unified in continuous representation, which justifies the potential effectiveness for HSIs.

    \item {We experimentally demonstrate that CLoRF significantly surpasses existing techniques, confirming its wide applicability and superiority and further expanding its application to HSI and PAN image fusion (HSI-PAN fusion).}
    
\end{enumerate}

The structure of the remainder of this paper is outlined as follows. Sec.~\ref{sec:related} provides a review of related work. Sec.~\ref{sec:method} introduces the proposed CLoRF in detail. Sec.~\ref{sec:experiment} presents the experimental results and subsequent discussion, and Sec.~\ref{sec:conclusion} concludes this work.

\section{Related Work}\label{sec:related}
\subsection{Implicit Neural Representation}

INR offers a novel approach to representing implicitly defined, continuous, differentiable signals parameterized by neural networks~\cite{sitzmann2020implicit}. INR has demonstrated remarkable performance in representing complex data structures, such as 3D reconstruction~\cite{mildenhall2021nerf,sitzmann2020implicit, takikawa2021neural} and 2D image super-resolution~\cite{dupont2021generative,anokhin2021image} and generation~\cite{xu2023revisiting,chen2021learning}, etc. Recently, INR-based approaches have been explored for HSIs, such as HSI SR~\cite{zhang2022implicit}, unmixing~\cite{wang2023nonnegative}, and fusion~\cite{wang2023ss, deng2023implicit}. Despite these commendable efforts, INR still encounters challenges in HSIs. For instance, HSIs consist of numerous spectral bands obtained through continuous imaging within a specific spectral range. However, INR itself may not possess sufficient stability to directly learn a valid continuous representation from the spectral domain of HSIs. In the fusion task, learning in both spatial and spectral domains is necessary; therefore, utilizing a single INR for learning representation may confront limited representation capabilities. 

\subsection{HSI-MSI Fusion via Continuous Representation}

Several recent studies explore INRs for HSI-MSI fusion. For instance, \citeauthor{wang2023ss} proposes spatial-INR and spectral-INR for spatial and spectral resolution reconstruction, respectively. Besides, \citeauthor{deng2023implicit} proposes an innovative fusion method that integrates CNN and INR. Based on them, \citeauthor{he2024two} develops two spectral-spatial INRs for arbitrary-resolution hyperspectral pansharpening. Although these fusion methods utilize spatial-spectral-based INRs, they significantly diverge from CLoRF. First, these methods rely on local implicit image functions \cite{chen2021learning} for HSI SR. They employ a complex CNN network to encode spatial and spectral features, then feed these features and the three-dimensional coordinates of HSIs into an MLP to recover HSIs. Second, these HSI-MSI fusion methods are supervised learning, which heavily relies on pairs of images for training. {Third, they do not perform low-rank matrix decomposition on HR-HSI, nor do they leverage the low-rank and smooth physical information inherent in HSIs. As a result, the computational cost of these INRs is high, primarily due to the large size of HSIs, which results in large three-dimensional coordinates.} Conversely, our method is model-based and unsupervised, finely encoding the low-rankness and smoothness into the continuous spatial-spectral factorization function. Consequently, this improves the stability of the continuous representation in HSIs and significantly reduces the complexity of the network structure and the computational cost. Therefore, our method is more feasible and general in various HSIs.

\section{Proposed Method}\label{sec:method}
In this section, we first present the problem formulation for HSI-MSI fusion, then introduce the details of our proposed framework in the following section.

\subsection{Problem Formulation}
Given the HR-MSI and LR-HSI data, we aim to approximate their corresponding HR-HSI data. Specifically, the HR-HSI, LR-HSI, and HR-MSI data are transformed into the matrix format along the spectral dimension. The HR-HSI is denoted as $\h Z\in \mathbb{R}^{L\times N}$, where $L$ is the number of spectral bands, and $N=H*W$ is the total number of pixels, in which $ H$ and $W$ indicate spatial resolution. The LR-HSI is denoted as $\h X \in \mathbb{R}^{L\times n}$, where $n$ represents the number of LR spatial pixels, i.e., $n \ll N$. Finally, the HR-MSI is denoted as $\h Y \in \mathbb{R}^{l\times N}$, where $l\ll L$ signifies that $\h Y$ has fewer spectral bands than $\h X$. 

The LR-HSI $\h X$ can be interpreted as a diminished-quality representation of HR-HSI  $\h Z$ in the spatial dimension, which is formulated as follows:
\begin{equation}\label{HSI}
    \h X=\h Z \h B \h S+ \h N_h, 
\end{equation}
where $\h N_h \sim \mathcal{N}(\boldsymbol{0}, \sigma_h \mathbf{I})$ represents the additive Gaussian noise. Besides, $\h B \in \mathbb{R}^{N\times N}$ is a spatial blurring operator of $\h Z$, representing the point spread function (PSF) of the hyperspectral sensor. Additionally, $\h S\in \mathbb{R}^{N\times n}$ is the spatial downsampling matrix.

Similarly, the HR-MSI $\h Y$ can be considered as a downsampled realization of HR-HSI $\h Z$ in the spectral dimension, which is formulated as:
\begin{equation}\label{MSI}
    \h Y=\h H\h Z+\h N_m, 
\end{equation}
where $\h H \in \mathbb{R}^{l\times L}$ is the spectral response function (SRF) and $\h N_m \sim \mathcal{N}(\boldsymbol{0}, \sigma_m \mathbf{I})$ denotes the additive Gaussian noise.

As HSIs generally have a low-rank structure, thus they lie in a low-dimensional subspace~\cite{simoes2014convex, zhuang2018fast}. The low-rank factorization aims to approximate a target matrix $\h Z$  as a product of two matrices:
\begin{equation}\label{NMF}
    \h Z \approx \h E \h A,
\end{equation}
where $\h E\in \mathbb{R}^{L\times K}$ is a spectral dictionary and $\h A\in \mathbb{R}^{K \times N}$  is a coefficient matrix,  respectively. And 
$K \ll L$ represents a hyperparameter controlling the number of spectral bases. The low-rank factorization representation offers three main advantages. First, it maximizes the utilization of strong correlations among the spectral bands. Second, by keeping $K$ small (where $K \ll L$), the size of the spectral mode is reduced, thereby enhancing computational efficiency. {Third, each column of matrix $\h Z$ can be linearly represented by the columns of matrix $\h E$ using the coefficients in matrix $\h A$. }
The rows of matrix $\h A$ maintain the spatial structures of matrix $\h Z$. {Note that \eqref{NMF} is not a unique factorization for $\h Z$. One could obtain another pairs $\hat{\h E} = \h E \h B$ and $\hat{\h A}= \h B^{-1} \h A$, with any inverse matrix $\h B \in \mathbb R^{K\times K}$.  }

By integrating Eq.\eqref{NMF} into Eq.\eqref{HSI} and Eq.\eqref{MSI}, $\h X$ and $\h Y$ are formulated as: 
\begin{equation}
    \h X=\h E \h A \h B \h S+\h N_h, \quad 
    \h Y=\h H \h E \h A+\h N_m.
\end{equation}

Thereafter, the fusion problem is transformed into the task of estimating the spectral dictionary $\h E$ and its corresponding coefficient matrix $\h A$ from matrices $\h X$ and $\h Y$, which follows the following optimization problem:
\begin{equation}\label{likelihood}
 \min_{\h E,\h A} \|\h X-\h E\h A\h B \h S\|_{\text{F}}^2+ \lambda \|\h Y-\h H \h E \h A\|_{\text{F}}^2,
\end{equation}
where $\Vert \cdot \Vert_{\text{F}}$ denotes the Frobenius norm, and $\lambda$ denotes the balancing factor. As there is a lack of specific prior information, the problem Eq.\eqref{likelihood} is undetermined; therefore, existing works focus on exploring the appropriate prior information. Nonetheless, these methods process HR-HSI within the dimensions of two modalities and fail to fuse the arbitrary resolutions of HSIs effectively.

\subsection{Deep Continuous Low-rank Factorization Model}

INR is widely adopted for learning continuous data representation, such as HSIs~\cite{wang2023ss,deng2023implicit}. However, simply utilizing a single INR to represent the HSI volume results in low efficiency and expensive computation, as it neglects the distinct low-rank structure of HSIs. Conversely, we propose a continuous low-rank factorization (CLoRF) model for learning HSI representation continuously and effectively. Our method fully explores the low-rank structure of HSIs by simultaneously learning its low-rank continuous representation by spatial and spectral INR. As a result, our approach effectively captures the low-rankness and smoothness of HSIs while overcoming the computational burden associated with existing  INR-based methods~\cite{wang2023ss,deng2023implicit} in HSIs. 


As illustrated in Fig.~\ref{flowchart}, we present an overview of the proposed CLoRF. Specifically, {CLoRF consists of two steps: low-rank decomposition and learning. The low-rank decomposition breaks down the HSI data space into two smaller subspaces: spatial basis $\h A$ and spectral transformation $\h E$.  Additionally, the spatial and spectral components are parameterized by two neural networks, using two INRs to learn the low-rank continuous representation of the HR-HSI.} 
Inspired by the advance of Sinusoidal Representation Networks (SIRENs)~\cite{sitzmann2020implicit}, we employ two SIRENs to estimate $\h E$  and $\h A$ in Eq.\eqref{NMF}, respectively. Specifically, we denote one SIREN $\Psi_{\theta}(\cdot)$ parameterize by $\theta$
for approximating $\h E$, and another SIREN $\Phi_{\alpha}(\cdot)$  parameterized by $\alpha$ for approximating $\h A$, which is defined as follows:
\begin{equation*}\label{coord}
    \begin{split}
    \h{\hat E}(\h b; \theta) & = [\Psi_{\theta}(b_1),\Psi_{\theta}(b_2),\dots, \Psi_{\theta}(b_L)]^T,  \\
    \h{\hat A} (\h O; \alpha) & = [\Phi_{\alpha}(\h o_{11}),\Phi_{\alpha}(\h o_{12}),\dots, \Phi_{\alpha}(\h o_{HW})],
    \end{split}
\end{equation*}
where 
$\Psi_{\theta}(b_i): \mathbb{R} \rightarrow \mathbb{R}^{K}$ is a spectral basis, with $b_i\in \mathbb{R}$  being the 1D coordinate for the  $i$-th band index of the LR-HSI. Besides,  $\Phi_{\alpha}(\h o_{ij}): \mathbb{R}^{2} \rightarrow \mathbb{R}^{K}$ is a spatial basis with the 2D coordinate $\h o_{ij} \in \mathbb{R}^ {2}$ of the  HR-MSI. And we denote the spectral bases as $\h b = [b_1,b_2, \dots, b_L]^T$ and the spatial bases as $\h O = [\h o_{11}; \h o_{12};\dots; \h o_{HW};]. $ Both networks aim to learn how to map from a fixed coordinate to the target representation. Here, we formalize these networks as follows: 
\begin{equation*}
\begin{aligned}
    &\Psi_{\theta}(b_i)=\h W^1_{d_1}(\cdots (\sigma(\h W^1_{1} b_i+\h c^1_1))\cdots)+\h c^1_{d_1}, \\
     &\Phi_{\alpha}(\h o_{ij})=\h W^2_{d_2}(\cdots(\sigma(\h W^2_{1} \h o_{ij}^T+\h c^2_1)\cdots)+\h c^2_{d_2}, \\
\end{aligned}
\end{equation*}
where $\sigma$ denotes the activation function, $\theta=(\{\h W^1_i\}_{i=1}^{d_1}, \{\h c^1_i\}_{i=1}^{d_1})$ and  
$\alpha=(\{\h W^2_i\}_{i=1}^{d_2}, \{\h c^2_i\}_{i=1}^{d_2})$ contains weight matrices and bias vectors for spectral- and spatial-INR, respectively.
Our method is a natural progression of low-rank factorization from discrete mesh grids to the continuous domain. And the target HR-HSI is approximated as $\h{\hat Z}= \h{\hat E} (\h b; \theta) \h{\hat A} (\h O; \alpha).$ 

As matrix $\h{\hat A} (\h O; \alpha)$ preserves the spatial structures of HSIs.  Without loss of generality, we consider the spatial smoothness of HSIs. Moreover, a total variation (TV) loss on the predicted coefficient matrix  $\h{\hat A} (\h O; \alpha)$ is further incorporated for noise-disruption scenarios. Mathematically, the TV regularization of $\h{\hat A} (\h O; \alpha)$ is formulated as:
\begin{equation}\label{tv_loss}
     \sum_{k=1}^K \text{TV}(\h {\hat a}_k)=\sum_{k=1}^K (\| \h D_h {\h {\hat a}}_k \|_1+ \| \h D_w {\h {\hat a}}_k \|_1),
\end{equation}
where $ \hat{\h a}_k$ is the $k$-th row of $\h{\hat A} (\h O; \alpha)$.
$\h D_h$ and $\h D_w$ denote the differential operation along the height and width direction in the matrix form of $ \hat{\h a}_k$, respectively.  Here, $ \|\cdot\|_1$ indicates the $\ell_1$ norm. By incorporating the TV loss, we promote spatial smoothness and improve
the overall quality of the fusion.   

Therefore, the optimization problem with TV  prior can be summarized
as follows:
\begin{equation}\label{eq:optim}
    \min_{\theta, \alpha} \mathcal{L}_{\text{MSI}}+ \lambda \mathcal{L}_{\text{HSI}}+\eta \sum\limits_{k=1}^K \text{TV}(\h {\hat a}_k),
\end{equation}
where $\mathcal{L}_{\text{MSI}}= \|\h X-\h{\hat E} (\h b; \theta) \h{\hat A} (\h O; \alpha)\h{BS}\|_\text{F}^2,$ $\mathcal{L}_{\text{HSI}}=\|\h Y-\h{H}\h{\hat E} (\h b; \theta) \h{\hat A} (\h O; \alpha)\|_\text{F}^2$, and  $\eta$ is the regularization parameter. 

Approximating $\h {\hat E}(\h b; \theta^\ast) $ and  $\h{\hat A} (\h O; \alpha^\ast) $ to maintain the low-rank representation of $\h{\hat Z}$ can be achieved after training networks, with $\theta^\ast, \alpha^\ast$ corresponding to the parameters of well-trained networks. Recall the networks take the coordinates as the input, and the optimization in Eq.\eqref{eq:optim} does not involve the ground-truth HSIs as the supervision label; therefore, our method is self-supervised. We employ the Adam optimizer for optimization, which is a stochastic gradient descent algorithm. Moreover, we can infer an arbitrary-resolution HSI by inputting any scale coordinates $\{\tilde{\h b}, \tilde{\h O}\}$ into the well-trained network, i.e., $\h{\hat E} (\tilde{\h b}; \theta^\ast) \h{\hat A} (\tilde{\h O}; \alpha^\ast)$.    

Compared to existing INRs-based fusion methods, our method stands out for several advantages. First, it fully exploits the low-rank and smooth prior of HSIs through low-rank continuous learning representations. Second, its computational complexity is significantly reduced through continuous low-rank factorization. Third, it can achieve the user-desired resolution at arbitrary locations in HSIs by inputting any scale spatial and spectral coordinates.

\subsection{Theoretical Analysis}
In this section, we theoretically demonstrate that the low-rank and smooth regularizations are implicitly unified in continuous low-rank matrix factorization. Our analysis is inspired by the concept of tensor function factorization in \cite{luo2023low}. Here, we start with rank factorization in the matrix computation field.

\begin{theorem}[rank factorization~\cite{piziak1999full}]\label{th1} 
Let $\h X \in \mathbb{R}^{n_1 \times n_2}$, where $\mathrm{rank} (\h X)=K$, then there exists two matrices $\h U \in \mathbb {R}^{n_1 \times K}$,  $\h V \in \mathbb {R}^{n_2 \times K}$ such that $\h X=\h U \h V^T$. 
\end{theorem}

Subsequently, we provide a detailed introduction to the proposed continuous representation of HSIs. Let $f(\cdot): \mathcal{A}_f \times \mathcal{Z}_f \rightarrow \mathbb {R}$ be a bounded real function, where $\mathcal{A}_f \subset \mathbb {R}^2$, $\mathcal{Z}_f \subset \mathbb {R} $ are definition domains in spatial and spectral domains, respectively. The function $f$ gives the value of data at any coordinate in $\mathcal{D}_f:= \mathcal{A}_f \times \mathcal{Z}_f$. We interpret $f$ as a matrix function since it maps a spatial and spectral coordinate to the corresponding value, implicitly representing matrix data.

\begin{definition}[sampled matrix set]
\label{def1}
     For a matrix function $f(\cdot):\mathcal{D}_f\to\mathbb{R}$, we define the sampled matrix set $\mathcal{S}[f]$ as
\begin{equation*}
    \begin{split}
        \mathcal{S}[f]:= \{ & \h M| \h M_{(i,j)}=f(\h s_i, b_j), \mathbf{s}_i \in \mathcal{A}_f, \\ & {b}_j \in \mathcal{Z}_f, 
         \h M \in \mathbb{R}^{n_1\times n_2}, n_1, n_2\in \mathbb N_+\},
    \end{split}
\end{equation*}
where $\h s_i, b_j$ denote the spatial and spectral coordinates, respectively. 
\end{definition}
Regarding the definition of rank, we expect any matrix sampled on $\mathcal{S}[f]$ to be low-rank. Naturally, we can then define the rank of the matrix function as follows.
\begin{definition}[matrix function rank]
\label{def2}
Given a matrix function $f: \mathcal{D}_f= \mathcal{A}_f \times \mathcal{Z}_f\to\mathbb{R}$, we define a measure of its complexity, denoted by $\mathrm{MF}\text{-}\mathrm{rank}[f]$ (function rank of  $f(\cdot) $), as the supremum of the matrix rank in the sampled matrix set $\mathcal{S}[f]{:}$
$$
\mathrm{MF} \text{-}\mathrm{rank}[f]:=\sup_{\h M \in \mathcal{S}[f]}\mathrm{rank}(\mathbf{M}). 
$$ 
\end{definition}

We call a matrix function $f(\cdot)$ as a low-rank matrix function if $K\ll \min \{n_1, n_2\}$. When $f(\cdot)$ is defined on a given matrix, we will show that the MF-rank degenerates into the discrete case, i.e., the classical matrix rank.

\begin{proposition}
\label{prop_1}
   Consider $\h {X}\in\mathbb{R}^{n_1\times n_2 }$ as an arbitrary matrix. 
   Let $\mathcal{A}_f= \mathcal{N}^{(l_1)} \times  \mathcal{N}^{(l_2)}  \ (l_1 l_2=n_1)$ represent a two-dimensional discrete set, and $\mathcal{Z}_f= \mathcal{N}^{(n_2)} $ is an one-dimensional discrete set, where $ \mathcal{N}^{(k)} $ is the set $\{1,2,\dots, k\}$. We denote $\mathcal{D}_f=\mathcal{A}_f\times \mathcal{Z}_f$ and define 
the matrix function $f(\cdot): \mathcal{D}_f\rightarrow \mathbb{R}$ as $f(\h s,b)=\h{X}_{(\h s,b)}$ for any $(\h s,b)\in \mathcal{D}_f$. Consequently, $\mathrm{MF}$-$\mathrm{rank}[f]=\mathrm{rank}(\h X)$.
\end{proposition}


%

The proof of Proposition~\ref{prop_1} is shown in Appendix~\ref{sec:proof_prop1}. Proposition~\ref{prop_1} expands the concept of rank from discrete matrices to matrix functions for continuous representations. Analogous to classical matrix representations, it is pertinent to consider whether a low-rank matrix function $f$ can employ certain matrix factorization strategies to encode its low-rank. We provide an affirmative response, as stated in the theorem below.
\begin{theorem} [continuous low-rank factorization]\label{thm_CLoRF}
 Let $f(\cdot):\mathcal{D}_f=\mathcal{A}_f\times \mathcal{Z}_f\to\mathbb{R}$ 
 be a bounded matrix function, where $\mathcal{A}_f\subset\mathbb{R}^2$, $\mathcal{Z}_f\subset \mathbb{R}$. 
    If $\mathrm{MF}$-$\mathrm{rank}[f]=K$, then there exist two functions $f_{\mathrm{spatial}}(\cdot): \mathcal{A}_f\rightarrow \mathbb{R}^K$, $f_{\mathrm{spectral}}(\cdot): \mathcal{Z}_f \rightarrow \mathbb{R}^K$ such that $f(\h s, b)= f_{\mathrm{spatial}}(\h s) \cdot f^T_{\mathrm{spectral}}(b)$ for any pair of inputs $(\mathbf{s}, b) \in \mathcal{D}_f$.
\end{theorem}
The proof of Theorem~\ref{thm_CLoRF} is illustrated in Appendix~\ref{sec:proof_thm3}. Theorem~\ref{thm_CLoRF} is a natural extension of rank factorization (Theorem~\ref{th1}) from discrete meshgrid to the continuous domain. Specifically, we employ two MLPs $\Phi_{\alpha}(\cdot)$ and $\Psi_{\theta}(\cdot)$ with parameters $\theta$ and $\alpha$ to parameterize the factor functions $f_{\mathrm{spatial}}(\cdot)$ and $f_{\mathrm{spectral}}(\cdot)$.
\begin{remark}
{1) In Theorem~\ref{th1}, the low-rank matrix decomposition definitely exists but is non-unique. This is because the representation of eigenvectors in SVD is not unique, which leads to different $\h U$ and $\h U'$, as well as $\h V$ and $\h V'$ in the decomposition, such as $\h U'=c \h U$, $\h V'=\h V/c$, where $c$ is a nonzero scalar.}

{2)  CLoRF comprises two subnetworks with identical architectures: Spatial-INR and Spectral-INR. This network framework, equipped with low-rank and smooth priors, learns to generate a spectral dictionary and spatial coefficient matrix, enabling a low-rank representation of HSIs through network training. The solution spaces of spatial and spectral basis are constrained by the parameters of the INRs. Furthermore, by incorporating TV prior into the loss function, the solution space of the spatial coefficient matrix is further restricted, effectively reducing the ambiguity in the low-rank factorization.}
\end{remark}

Smoothness is another prevalent attribute in HSIs, such as the spatial and spectral smoothness of HSIs~\cite{sun2021deep}. Here, we theoretically validate that our method incorporates implicit smooth regularization derived from the specific structures of MLPs.
\begin{theorem}[Lipschitz continuity]
\label{lips}
    Let $\h X \in \mathbb{R}^{n_1\times n_2}$, and  $\Phi_{\alpha}(\cdot): \mathcal{A}_f \rightarrow \mathbb{R}^K$, $\Psi_{\theta}(\cdot): \mathcal{D}_f \rightarrow \mathbb{R}^K$ be two MLPs structured with  parameters $\alpha, \theta$ where $\mathcal{A}_f \subset \mathbb{R}^2$, $\mathcal{Z}_f \subset \mathbb{R}$ are bounded, i.e., $\|\h s\|_1\leq \zeta, $  $b \leq \zeta$ for any $\h s\in \mathcal{A}_f$, $ b\in \mathcal{Z}_f$.  Suppose the MLPs share the same activation function $\sigma(\cdot)$ and depth $d$ with $\h c^1_i=\h c^2_i = \h 0, \forall \ i$.  Besides, we assume that
\begin{itemize}
    \item $\sigma$ is Lipschitz continuous with the Lipschitz  constant $\kappa$, and $\sigma(0)=0$;
    \item $\|\h W_i^1\|_1, \|\h W_i^2\|_1$ are bounded by a positive constant {$\eta$} for all $i$.
\end{itemize}
Define a matrix function $f(\cdot): \mathcal{D}_f=\mathcal{A}_f\times \mathcal{Z}_f \rightarrow \mathbb{R}$ as $f(\h s, b)=\Phi_{\alpha}(\h s)\cdot \Psi_{\theta}(b)^T$. Then, the following inequalities hold for any $(\h s_1,b_1), (\h s_2, b_2)\in \mathcal{D}_f$:
\begin{equation*}
\begin{aligned}
    |f(\h s_1,b_1)-f(\h s_2,b_2)| & \leq \delta \|\h s_1-\h s_2\|_{1} +  \delta |b_1-b_2|,
\end{aligned} 
\end{equation*}
where $\delta=\eta^{2d+1} 
\kappa ^{2d-2}\zeta$, 
and {$\zeta=\max\{\|\h s_1\|_{1},|b_1|\}$}. 
\end{theorem}


The proof is shown in Appendix~\ref{sec:proof_thm4}. This smoothness is implicitly encoded with mild assumptions regarding nonlinear activation functions and weight matrices, which are readily attainable in a real-world implementation. For instance, {we utilized the sine activation function, while also ensuring that the weights in the MLP network~\cite{sitzmann2020implicit} remain bounded. }

\begin{remark}
{In Theorem~\ref{lips}, we observe that the degree of smoothness, denoted as $\delta$, is associated with the Lipschitz constant $\kappa$ and the upper bound $\eta$ of the weight matrices. Therefore, we can manipulate two variables in practice to achieve a balance in implicit smoothness:}

{
1) We utilize the Sine function $\sigma(\cdot) = \sin(\omega_0 \cdot)$ as the nonlinear activation function in the MLPs. Since the Sine function is Lipschitz continuous, we can effectively adjust its Lipschitz constant $\kappa$ by varying the value of $\omega_0$. Specifically, a smaller $\omega_0$ results in a lower Lipschitz constant $\kappa$, leading to smoother outcomes.}

{2) To manage the upper bound $\eta$ of the MLP weight matrices, we can adjust the trade-off parameter in the energy regularization of the MLP weights, commonly referred to as weight decay in contemporary deep learning optimizers. This approach allows us to control the strength of $\eta$.}
\end{remark}

\begin{remark}
\label{re2}
{Assume the assumptions in Theorem~\ref{lips} are satisfied. We define $f(\cdot):=[\Phi_{\alpha},\Psi_{\theta}] (\cdot)$. Then, for any matrix $\h M \in \mathcal{S}[f]$ that is  sampled using coordinates vectors $\h s \in\mathcal{A}_f$,  $t \in \mathcal{Z}_f$, where  $ \mathcal{S}[f]$ represents the set of sampled matrices from the matrix function $f(\cdot)$ as defined in Definition~\ref{def1}, the following inequalities hold for  $(i,j), i=1,2,\cdots,n_1, j=1,2,\cdots n_2:$
\begin{equation}
\small
     |\h M_{(\h s_i,t_j)}-\h M_{(\h s_{i-1},t_{j-1})}| 
 \leq \delta \|\h s_i-\h s_{i-1}\|_1 + \delta |t_j-t_{j-1}|,
\end{equation}
where $\delta=\eta^{2d+1} \kappa ^{2d-2}\zeta$, and $\zeta=\max\{\|\h s_1\|_{1},|b_1|\}$.}
\end{remark}
{Remark \ref{re2} states that for any sampled matrix $\h M \in S[f]$,
the difference between adjacent elements is constrained by the distance between the corresponding coordinates, with the inclusion of a constant factor.}

\section{Experiments and Analysis}\label{sec:experiment}

In this section, we evaluate the performance of our method on five datasets separately. Additionally, we compare several SOTA methods and evaluate the fusion results qualitatively and quantitatively. Finally, we expand the application of CLoRF to HSI-PAN fusion and verify its efficacy on a dataset.

\subsection{Experimental Details}\label{experiment_detail}

1) Datasets: 
{We evaluate the performance of fusion using both synthetic and real datasets. }
{Seven} simulated datasets, including Pavia University~\cite{dell2004exploiting}, Pavia Center~\cite{dell2004exploiting}, Indian Pines~\cite{baumgardner2015220}, Washington DC~\cite{zhuang2021fasthymix, zhuang2018fast}, University of Houston~\cite{le20182018}, {Peppers~\cite{yasuma2010generalized}, and Superballs~\cite{yasuma2010generalized} }were used for simulations in our experiments.
The {seven} synthetic datasets, each with a simulated PSF and SRF, are used to generate two observed images: LR-HSI and HR-MSI. More specifically, we utilize a Gaussian blur of $5\times5$ pixels with a 0 mean and 1 standard deviation to simulate the PSF for generating the LR-HSI. The downsampling ratio was set to 4 for all datasets. The spectral response of the IKONOS satellite (a Nikon D700 camera) was used to simulate the SRF for generating the HR-MSI. The i.i.d Gaussian noise was added to HR-MSI and LR-HSI with signal-to-noise ratios (SNRs) of 30 dB, respectively. The details for the {eight} datasets are summarized as follows. 

\begin{itemize}

\item Pavia University: The image measures $610\times340\times115$ pixels with a spatial resolution of 1.3 meters and spectral coverage spanning from 0.43 $\mu m$ to 0.86 $\mu m$. Due to the effects of noise and water vapor absorption, 12 bands were removed. An area covering $336 \times 336$ pixels in the lower-left corner of the image and containing 103 bands was selected for the experiment.

\item Pavia Center:  The size of the Pavia Center is $1096\times1096\times115$  and spectral ranging from 0.38 to 1.05 $\mu m$. After removing bands caused by water vapor absorption and low SNRs, the subregion consisting of $336 \times
336 \times 93$ pixels were chosen from the full dataset as a reference HR-HSI. 

\item Indian Pines: The Indian Pines has 224 spectral bands with a size of $145 \times145$ pixels.  the spectral wavelength range is from 0.4 to 2.5 $\mu m$.
After removing bands caused by water vapor absorption and low SNRs, the subregion consisting of $144 \times 144 \times 191 $ pixels was chosen from the full dataset as a reference HR-HSI.

\item Washington DC Mall: The Washington DC Mall has a region of $1280\times307$ pixels, and the image consists of 210 bands with spectral wavelength ranging from 0.4 to 2.5 $\mu m$. After removing the low SNRs and atmospheric absorption bands, 191 bands were kept. The subregion consisting of $304\times304 \times191$ pixels was clipped from the full dataset as a reference HR-HSI.

\item University of Houston:  The  University of Houston contains $601\times2384$ pixels and 48 bands ranging from 0.38 to 1.05 $\mu m$. The sub-image consisting of $320\times320\times46$ pixels was chosen from the whole dataset as a reference HR-HSI for our experiments.

\item {Peppers and Superballs}:   \tb {We utilize the widely-used CAVE dataset \cite{yasuma2010generalized}, which is a ground-based HSI dataset commonly employed in HSI-MSI fusion.  The dataset comprises 32 high-resolution HSIs, each with a size of $512 \times 512 \times 31$. 
}

\item {Real Data: The LR-HSI is collected by the Hyperion sensor~\cite{yang2018hyperspectral}, which is of the size $120\times 120 \times 89$. The HR-MSI with 13 bands is taken by the Sentinel-2A satellite. Four bands are employed for the test, and the spatial downsampling factor is 3, i.e., the size of
 HR-MSI is $360 \times 360 \times 4$.  These four bands are extracted from bands 2, 3, 4, and 8, with the central wavelengths being  490, 560, 665, and 842  nm, respectively. }
\end{itemize}

2) Comparison methods: Since our method is an unsupervised algorithm, we mainly compare it to unsupervised fusion methods; such a choice also aligns with the practical demands of real-world scenarios. Specifically, we compare the proposed method with several SOTA unsupervised fusion methods, including
CNMF~\cite{yokoya2011coupled}, HySure~\cite{simoes2014convex}, HyCoNet~\cite{zheng2020coupled}, 
CNN-FUS~\cite{dian2020regularizing}, MSE-SURE~\cite{nguyen2022deep}, and E2E-FUS~\cite{wang2023self}. For baselines: CNMF, HySure, CNN-FUS, and E2E-FUS methods are implemented using MATLAB R2023a on Windows 11 with 32GB RAM, while HyCoNet, MSE-SURE, CLoRF, {and supervised methods} are evaluated on an RTX 4090 GPU with 32GB RAM. Besides, parameters in baselines are manually fine-tuned to achieve optimal results in all experiments.

3) Evaluation metrics: Four distinct quantitative metrics are employed to assess the efficacy of fusion results,  where ground truth (GT) is given. These metrics include the mean structured similarity (MSSIM), an extension of SSIM to assess HSI quality by averaging across all spectral bands; mean peak signal-to-noise ratio (MPSNR), which is calculated as the average PSNR across all bands extended for HSI; the spectral angle mapper (SAM) index; and the relative global dimension error (ERGAS) index.

4) Hyperparameters of CLoRF:
For all tasks, both the spatial-INR and spectral-INR adopt the SIREN network with an initial network parameter $\omega_0=30$. The spatial-INR has 5 hidden layers with a size of 512, while the spectral-INR has 2 hidden layers with a size of 128 (In Indian Pines, the spatial-MLP has 3 hidden layers with a size of 512, while the spectral-MLP also has 3 hidden layers with a size of 256). {Spatial-INR and spectral-INR are jointly optimized based on the loss function \ref{eq:optim}, using the Adam optimizer for optimization.}
The learning rate is set to 3e-5, and the training epochs are fixed to 30000. An early stopping strategy is implemented to prevent overfitting for all datasets. Furthermore, the hyperparameters are summarized in Table~\ref{hyperpara}.

\begin{table*}[t!]
  \centering
  \caption{Hyperparameters used for training the proposed model.}
\footnotesize
  \begin{tabular}{cccccccc}
    \toprule
\label{hyperpara}
    Hyperparameter& Pavia University & Pavia Center & Indian Pines & Washington DC  &  Houston & {Peppers}& {Superballs}\\
    \midrule
     $K$  &9 & 9 & 22 & 11    &   10   &   10 &   10    \\
    $\lambda$ & 1.25 & 1.25 & 0.55 &1.25 &  1 &   1 &   1\\ 
    $\eta$ & 0.0025 & 0.0050 & 0.0060 &0.0025  & 0.0025 &   0.0025 &   0.0025\\
    \bottomrule
  \end{tabular}
  \label{hyper}
\end{table*} 

\begin{table*}[t!]
   \centering
       \caption{Quantitative performance comparison with different algorithms on the different datasets. The best results are \textbf{bold-faced}, and runner-ups are \underline{underlined}. (MPSNR \textuparrow, MSSIM \textuparrow, SAM \textdownarrow, ERGAS \textdownarrow). }
    \label{all_result}
    \footnotesize
    \begin{tabular}{ccccccccc}
      \toprule
      &  Metric & CNMF &  HySure      &HyCoNet & CNN-FUS &MSE-SURE  &E2E-FUS & CLoRF  \\   \midrule
   \multirow{4}*{Pavia University}   
       & MPSNR & 32.70 & 32.86 & 40.10 & \underline{41.57} & 41.31 & 40.92 & \textbf{42.23} \\
         & MSSIM & 0.92 & 0.93 & 0.97 & \underline{0.98} & \underline{0.98} & \underline{0.98} & \textbf{0.99} \\
         & SAM & 3.55 & 5.81 & 3.10 & 2.38 & \underline{2.20} & 2.31 & \textbf{2.05} \\
        & ERGAS & 3.63 & 3.71 & 1.41 & \underline{1.38} & 1.44 & 1.50 & \textbf{1.31} \\
              \midrule
    \multirow{4}*{Pavia Center}   
       & MPSNR & 38.97 & 33.96 & 41.88 & 42.16 & 42.87 & \underline{43.12} & \textbf{43.47} \\
         & MSSIM & 0.97 & 0.93 & \underline{0.98} & \underline{0.98} & \textbf{0.99} & \textbf{0.99} & \textbf{0.99} \\
         & SAM & 7.70 & 8.42 & 3.77  & 4.10 & \textbf{3.51} & 3.67 & \underline{3.64} \\
        & ERGAS & 2.35 & 3.56 & 1.45  & 1.47 & 1.37 & \underline{1.33} & \textbf{1.26} \\
        \midrule
     \multirow{4}*{Indian Pines}  
      & MPSNR & 27.48 & 26.54 & 29.44 & 31.07 & 30.85 & \underline{31.68} & \textbf{31.96} \\
        &  MSSIM & \underline{0.94} & 0.93 & 0.93 & \underline{0.94} & \textbf{0.95} & \textbf{0.95} & \textbf{0.95} \\
          & SAM & 3.11 & 3.97 & 2.87 & 2.70 & \underline{2.54} & \textbf{2.49} & 2.56 \\
          & ERGAS & 2.52 &2.84 & 2.10 & 1.78 & 1.93  & \underline{1.73} & \textbf{1.70} \\    \midrule
           \multirow{4}*{Washington DC}  
      & MPSNR & 28.47 & 26.37 & 33.35 & 32.24 & 33.52 & \underline{36.15} & \textbf{36.75} \\
        & MSSIM & 0.94 & 0.94  & \underline{0.98} & 0.96 & \underline{0.98} & \underline{0.98} & \textbf{0.99} \\
          & SAM & 4.94 & 7.67 & 3.51  & 4.07 & \textbf{2.56} & \textbf{2.56} & \underline{2.64} \\
          &  ERGAS & 3.22 & 4.79  & 2.67 & 2.37 & 1.98 & \underline{1.53} & \textbf{1.44} \\    \midrule
    \multirow{4}*{University of Houston}  
      & MPSNR & 31.71 & 29.41  & 37.02 & 37.38 & \underline{39.75} & 38.84 & \textbf{40.55} \\
        & MSSIM & 0.95 & 0.92 & 0.97 & 0.97 & \textbf{0.99} & \underline{0.98} & \textbf{0.99} \\
          & SAM & 2.77 & 4.77 & 2.50 & 2.46 & \underline{1.61} & 1.76 & \textbf{1.47} \\
          & ERGAS & 2.10 & 3.13 & 1.19 & 1.31 & \underline{0.90} & 1.06 & \textbf{0.83}\\ \midrule
    \multirow{4}*{ {Peppers}}
      &   MPSNR &   40.00 &   36.21 &   41.15 &   \underline{44.86} &   43.25 &   43.23 &   \textbf{47.65} \\
        &   MSSIM &   0.96 &   0.94 &   0.96 &   \textbf{0.99} &   \textbf{0.99} &   \underline{0.98} &   \textbf{0.99} \\
          &  SAM &   11.62 &   10.59 &   \underline{4.77} &  7.36 &   6.34 &   7.02 &   \textbf{4.01} \\
          &   ERGAS &   4.92 &   9.04 &   6.29 &   \underline{2.65} &  2.87 &   2.94 &   \textbf{1.94}\\
          \midrule
    \multirow{4}*{ {Superballs}  }
     &    MPSNR &   43.88  &   39.14&   43.05 &   45.26 &   44.74 &   \underline{45.80} &    \textbf{46.14} \\
        &   MSSIM &   0.97&      0.96 &   \underline{0.98} &   \underline{0.98} &    \underline{0.98}   &   \underline{0.98} &   \textbf{0.98}   \\
          &   SAM &   10.84 &   10.40 &  7.98 &  7.47 &    \underline{6.84} &   7.42 &   \textbf{6.40} \\
          &   ERGAS &   3.58 &   5.82 &   5.98 &   2.92 &    3.10&   \underline{2.72} &    \textbf{2.71}  \\
          \bottomrule
    \end{tabular}
\end{table*}

\begin{figure*}[t!]
\begin{center}
\includegraphics[width=1\textwidth]{ 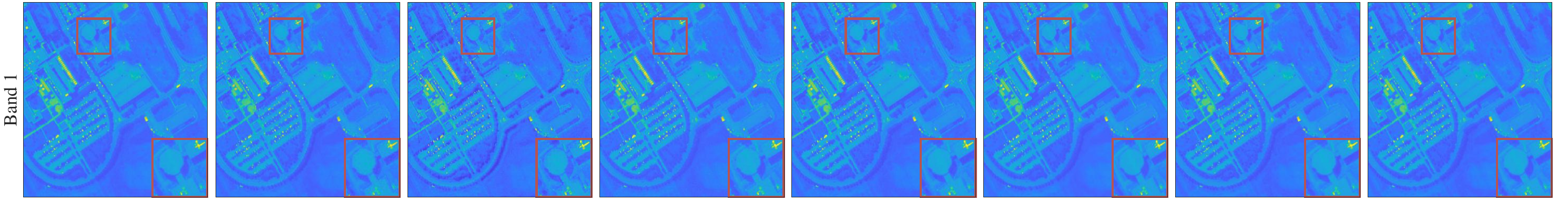} 
\includegraphics[width=1\textwidth]{ 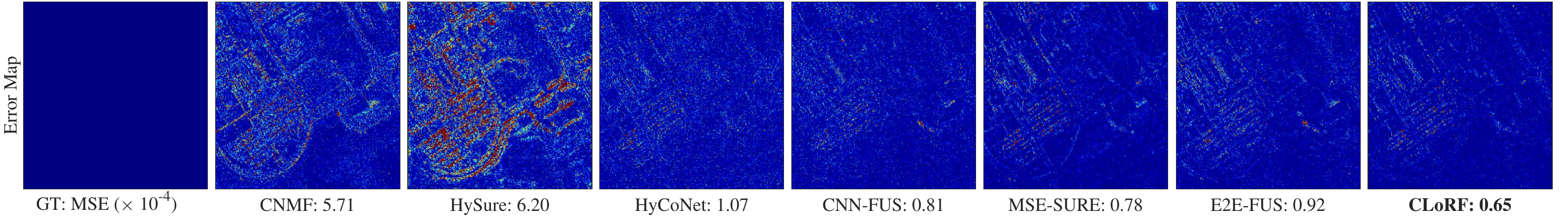} 
\end{center}
\caption{The first row shows the Pavia University image (1st band) of the estimated HR-HSI, and the second row shows the error map between the estimated image and GT.}
\label{pav}
\end{figure*}
\begin{figure*}[t!]
\begin{center}
\includegraphics[width=1\textwidth]{ 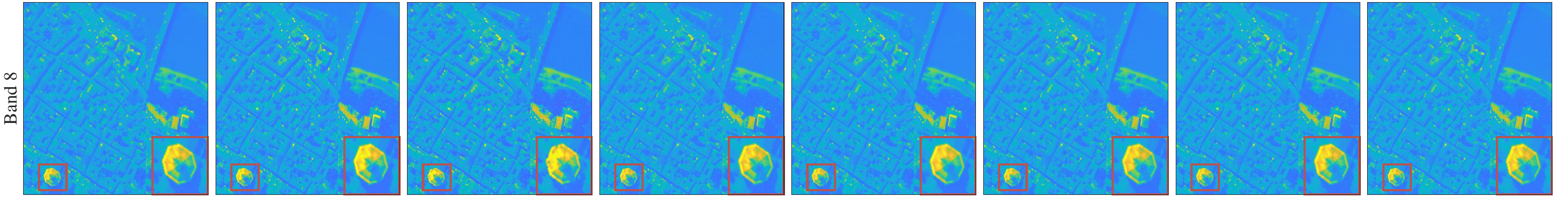} 
\includegraphics[width=1\textwidth]{ 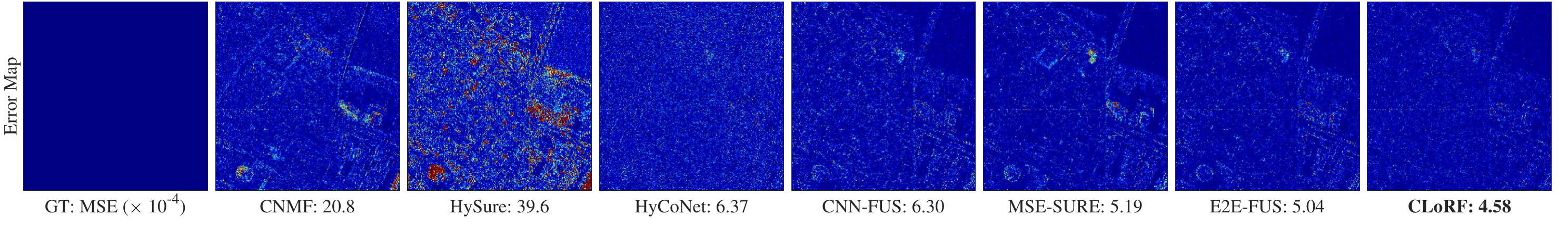} 
\end{center}
\caption{The first row shows the Pavia Center image (8th band) of the estimated HR-HSI, and the second row shows the error map between the estimated image and GT.}
\label{pav_center}
\end{figure*}
\begin{figure*}[t!]
\begin{center}
\includegraphics[width=1\textwidth]{ 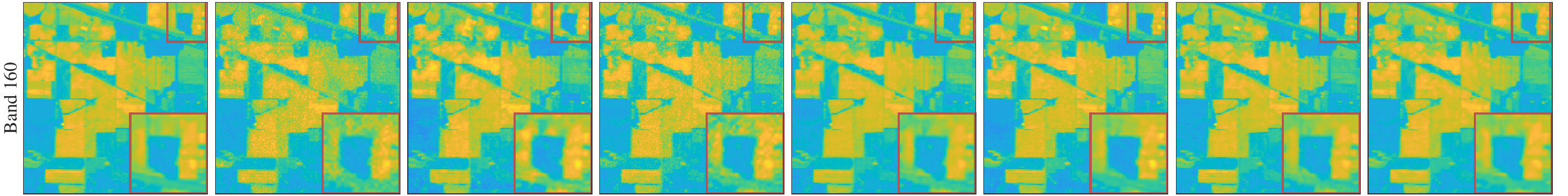} 
\includegraphics[width=1\textwidth]{ 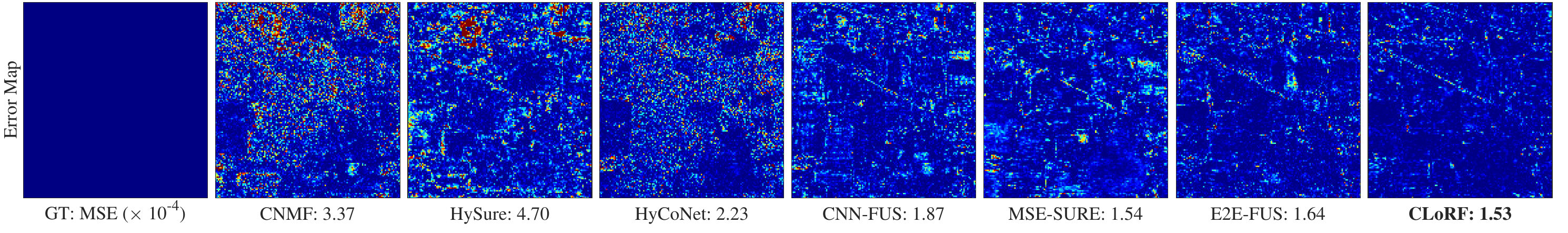} 
\end{center}
\caption{The first row shows the Indian Pines image (160th band) of the estimated HR-HSI, and the second
row shows the error map between the estimated image and GT. }
\label{Indian}
\end{figure*}
\begin{figure*}[t!]
\begin{center}
\includegraphics[width=1\textwidth]{ 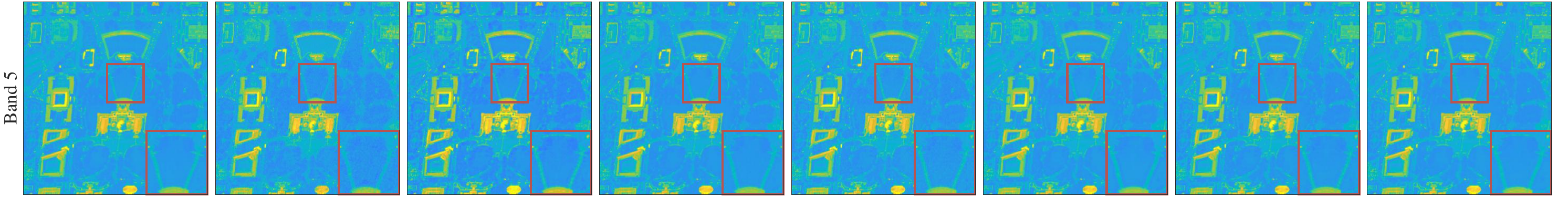} 
\includegraphics[width=1\textwidth]{ 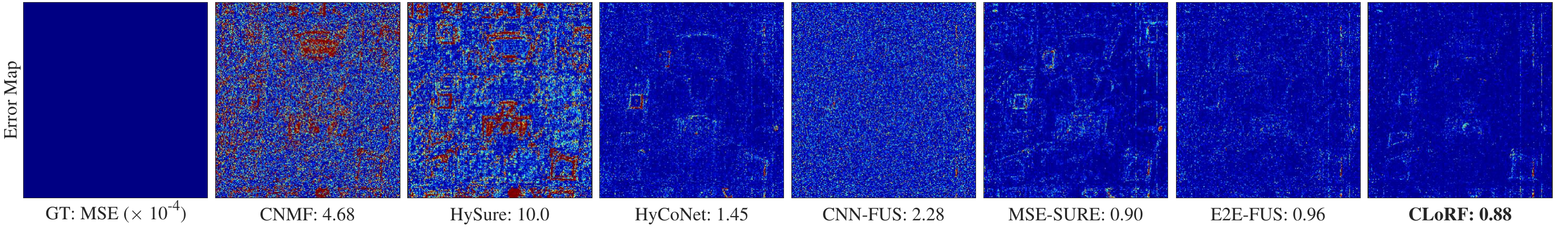} 
\end{center}
\caption{The first row shows the Washington DC image (5th band) of the estimated HR-HSI, and the second
row shows the error map between the estimated image and GT. }
\label{Wash_dc}
\end{figure*}
\begin{figure*}[t!]
\begin{center}
\includegraphics[width=1\textwidth]{ 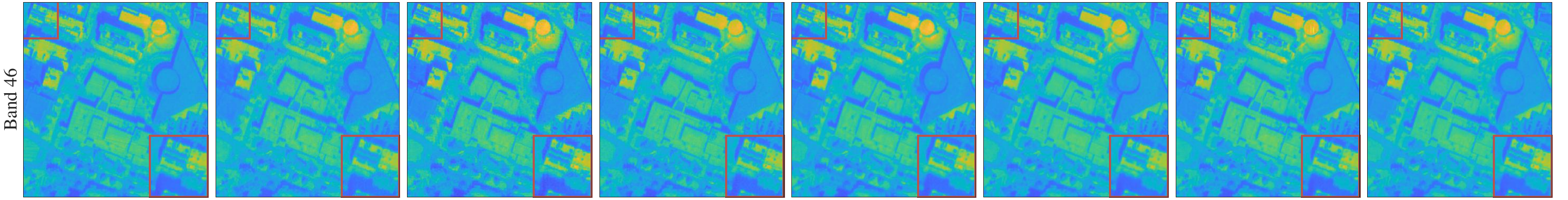} 
\includegraphics[width=1.001\textwidth]{ 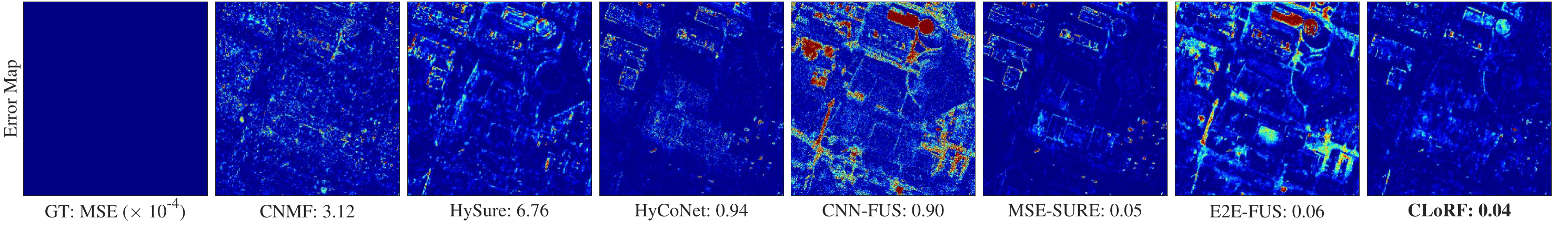} 
\end{center}
\caption{The first row shows the University of Houston image (46th band) of the estimated HR-HSI, and the second
row shows the error map between the estimated image and GT. }
\label{houston}
\end{figure*}
\begin{figure*}[t!]
\begin{center}
\includegraphics[width=1\textwidth]{ 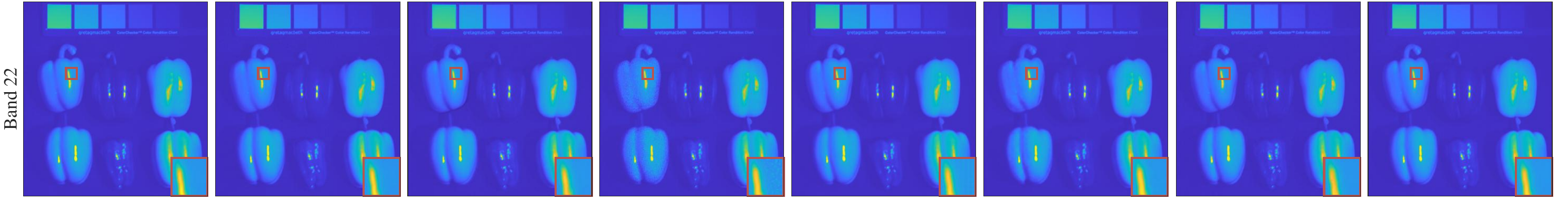} 
\includegraphics[width=1.001\textwidth]{ 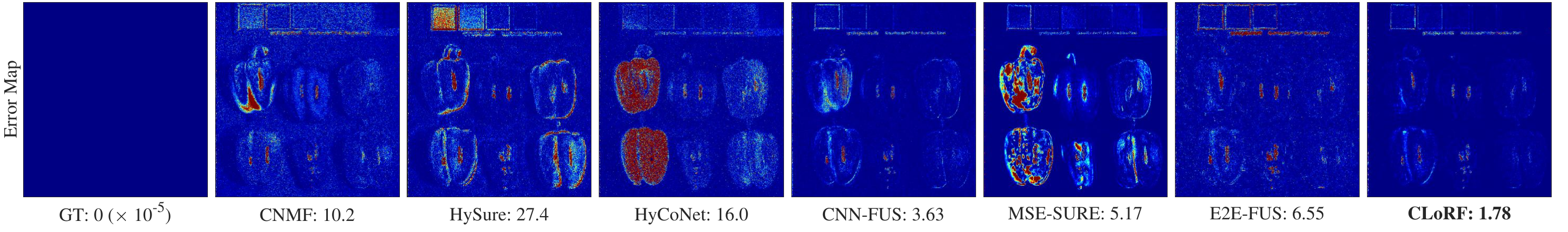} 
\end{center}
\caption{{The first row shows the Peppers image (22th band) of the estimated HR-HSI, and the second
row shows the error map between the estimated image and GT.} }
\label{peppers}
\end{figure*}
\begin{figure*}[t!]
\begin{center}
\includegraphics[width=1\textwidth]{ 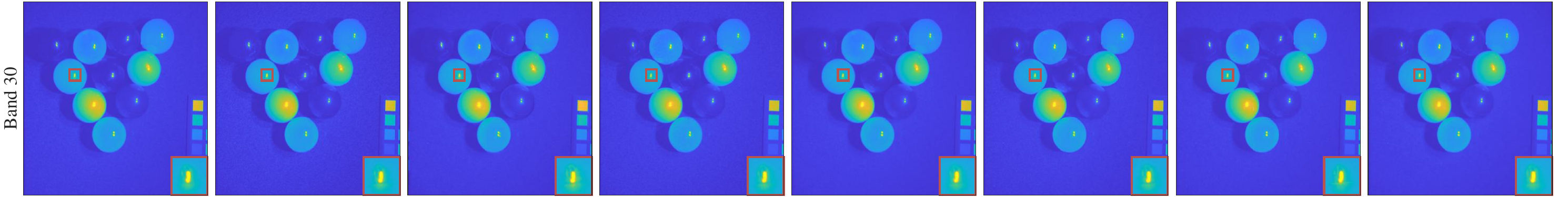} 
\includegraphics[width=1.001\textwidth]{ 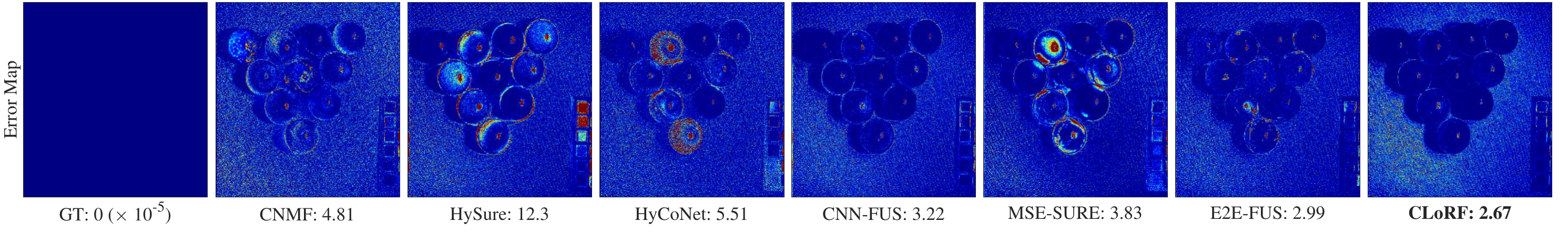} 
\end{center}
\caption{ {The first row shows the Superballs image (30th band) of the estimated HR-HSI, and the second
row shows the error map between the estimated image and GT.} }
\label{superball}
\end{figure*}
\begin{figure*}[t!]
\begin{center}
\includegraphics[width=1\textwidth]{ 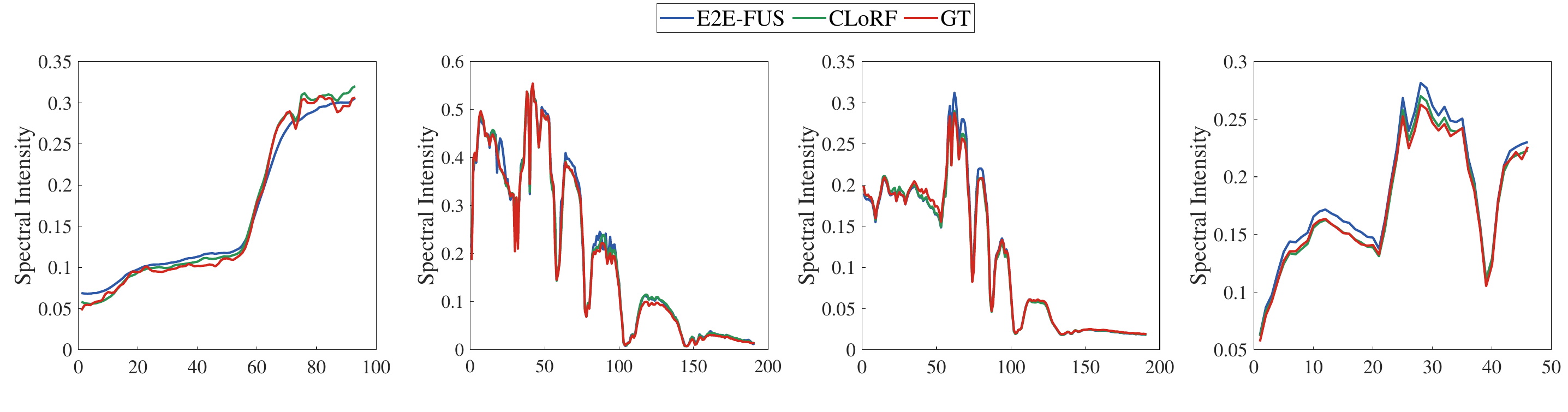} 
\includegraphics[width=1\textwidth]{ 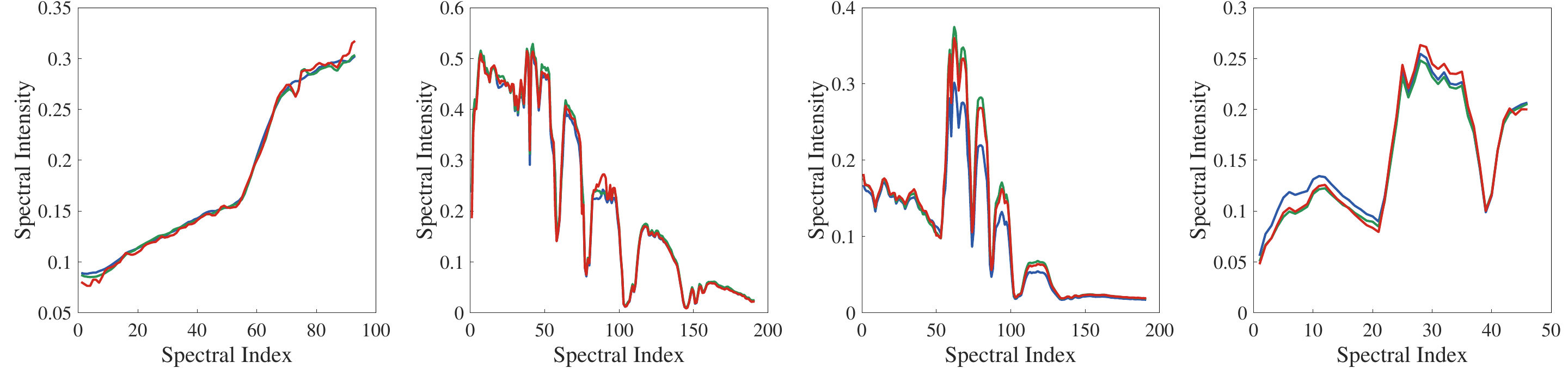} 
\caption{Reconstructed spectral signatures of two randomly selected locations at different datasets. 
From left to right, they are Pavia University, Indian Pines, Washington DC, and University of Houston. To provide a clearer comparison, we have compared CLoRF alongside E2E-FUS and GT. }
\label{different_pixel}
\end{center}
\end{figure*}
\subsection{Evaluation on Synthetic Data}
{Here, we consider synthetic data and evaluate proposed CLoRF under various scenarios. }

1) HSI-MSI fusion: 
The fusion performance of all methods on seven datasets, evaluated across MPSNR, MSSIM, SAM, and ERGAS metrics, is shown in Table \ref{all_result}. CLoRF consistently outperforms others in terms of quantitative results across most scenarios, affirming the efficacy of the HSI-MSI fusion task. Moreover, CLoRF exhibits commendable performance in spatial structures. our method outperforms E2E-FUS by 2.19\% in terms of MPSNR. Nonetheless, in certain datasets, particularly those with a high number of bands, CLoRF slightly lags behind MSE-SURE in SAM index performance. 
 It is notable that  MSE-SURE, throughout its training phase, is equipped with knowledge of noise levels in LR-HSI and HR-MSI, and leverages back-projection techniques to enhance spectral detail capture, while our method does not require this information. 
 
To assess the effectiveness of various methods in preserving spatial structures, {Fig.~\ref{pav}-\ref{superball} illustrates estimated HR-HSI on seven simulated datasets.} The first rows exhibit fusion results virtually, showcasing the 1st band of Pavia University, the 8th band of Pavia Center, the 160th band of Indian Pines, the 5th band of Washington DC Mall, the 46th band of the University of Houston, {the 22nd band of the Peppers, and the 30th band of the Superballs}. Conversely, the second rows depict corresponding error maps, illustrating the mean squared error (MSE) between GT and the estimated HR-HSI. While the results for the images in the first rows appear similar across most methods, with differences almost imperceptible to the naked eye, notable disparities emerge in the error maps depicted in the second column. It is evident that CLoRF exhibits minimal noise and preserves superior spatial structures. As shown in Fig.~\ref{different_pixel}, we plot the two randomly selected spectral vectors reconstructed by CLoRF and E2E-FUS with four datasets. It is apparent that CLoRF effectively preserves the high-frequency information compared to E2E-FUS in some pixels. However, in some specific pixels, the performance of the E2E-FUS method is superior to CLoRF.

{2)  Runtime and complexity analysis:  We have conducted a runtime and complexity analysis of various methods on the Houston dataset, with the experimental results reported in Table~\ref{running_time}. Please note that the runtime for supervised learning methods includes both training and inference time. Plug-and-play methods require pre-trained networks. In comparison to self-supervised learning methods, CLoRF demonstrates a significant advantage in terms of runtime efficiency and GFLOPs. Additionally, CLoRF excels in running time and model parameters when compared to supervised learning baselines. }
\begin{table}[t!]
   \centering
    \caption{{Runtime, parameters, and GFLOPs of all methods for Houston dataset. The best results are \textbf{bold-faced}.}}
     \footnotesize
    \tabcolsep=0.01cm
    \label{running_time}
    {
    \begin{tabular}{ccccc}
      \toprule
       & Methods & Times (s)  &   Paras (M)  & GFLOPs   \\   \midrule 
    \multirow{2}*{Model-based}     & CNMF  &    \textbf{5.70} (CPU)    &    /     &      /  \\
      &  HySure &   57.60 (CPU)    &   /      &       /   \\
      \midrule
      \multirow{2}*{Plug-and-Play}
       & CNN-FUS  &   114.06 (CPU)  &      /  &       /  \\
          &  E2E-FUS   &   {\bf 107.00}  (CPU)    &  /   &    / \\  \midrule
      \multirow{3}*{Deep-learning}     &  HyCoNet &   1332.38  (GPU)  &      {\bf 0.58}    &     152.63        \\
      &   MSE-SURE &   478.23  (GPU)&    1.10   &   149.92       \\
     (Self-supervised)  &  CLoRF   & \textbf{446.86} (GPU)  &  1.36  &  \textbf{121.75}     \\ \midrule
   \multirow{3}*{Deep-learning}     &  ConSSFCNN &   {\bf 473.51} (GPU)  &      4.79   &     78.37      \\
      &   ResTFNet &   525.07 (GPU)&   23.24  &  84.73      \\
       (Supervised)     &   SSR-Net &   504.51 (GPU)&    {\bf 2.87}  &   {\bf 46.69}       \\
   &  MCT-Net  &   1128.71 (GPU)  &  36.09  & 117.86    \\ 
          \bottomrule
    \end{tabular}
    }
\end{table}

{3) Performance on different downsampling ratios: 
We evaluate the model's performance when there is a significant difference in resolution between the input LR-HSI and the desired HR-HSI. Specifically, we choose the downsampling ratio in $\{ 4, 8, 16 \}$. The experimental results are shown in Table~\ref{diff_scale}, where we observe that our method still demonstrates its advantage when there is a large resolution gap between the LR-HSI and HR-HSI.}
\begin{table}[t!]
   \centering
    \caption{{Performance on different downsampling ratios with 
 30 dB noise for Pavia University. The best results are \textbf{bold-faced}.}}
    \label{diff_scale}
    \footnotesize
    {
    \begin{tabular}{cccccc}
      \toprule
     Ratios &  Methods &  MPSNR &  MSSIM  &SAM &  ERGAS   \\   \midrule
        \multirow{7}*{4}   
       & CNMF & 32.70 & 0.92 & 3.55& 3.63\\
         & HySure & 32.86 & 0.93 & 5.81 & 3.71 \\
         & HyCoNet& 40.10 & 0.97 & 3.10& 1.41  \\
        & CNN-FUS& 41.57& 0.98 & 2.38& 1.38  \\  
          & MSE-SURE& 41.31 & 0.98 & 2.20& 1.44 \\
        & E2E-FUS& 40.92 & 0.98 & 2.31& 1.50  \\  
          & CLoRF& \textbf{42.23} & \textbf{0.99}& \textbf{2.05}& \textbf{1.31} \\  
        \midrule
   \multirow{7}*{8}   
        & CNMF & 24.23 & 0.74& 7.67& 4.82\\
         & HySure & 23.87 & 0.74 & 12.55 & 4.88 \\
         & HyCoNet& 39.67 & 0.97 & 3.18& 0.73  \\
        & CNN-FUS& 40.09& 0.97 & 2.76& 0.85 \\  
          & MSE-SURE& 40.71 & \textbf{0.98} & 2.40& 0.87\\
        & E2E-FUS& 40.29 & \textbf{0.98} & 2.61& 0.83 \\  
          & CLoRF& \textbf{41.56} & \textbf{0.98}& \textbf{2.29}& \textbf{0.69} \\  
              \midrule
    \multirow{7}*{16}   
         & CNMF & 21.27 & 0.62& 12.32&3.50\\
         & HySure & 17.54 & 0.58 & 21.06& 4.67\\
         & HyCoNet& 39.47 & 0.97 & 3.25& 0.37  \\
        & CNN-FUS& 40.08& \textbf{0.98}& 2.70& 0.42 \\  
          & MSE-SURE&39.63& \textbf{0.98} & 2.92& 0.44\\
        & E2E-FUS& 40.21 & \textbf{0.98} & 2.63& 0.42\\  
          & CLoRF& \textbf{41.11} & \textbf{0.98}& \textbf{2.33}& \textbf{0.37} \\  
          \bottomrule       
    \end{tabular}}
\end{table}

{4) Comparison with supervised methods: To provide a broader context for the performance of the proposed approach, we compare it with recent four supervised methods, such as ConSSFCNN \cite{han2018ssf}, ResTFNet \cite{liu2020remote}, SSR-Net \cite{zhang2020ssr}, MCT-Net \cite{wang2023mct}. For fairness, since our approach is self-supervised, we evaluate it on the same test sets of the datasets used by the supervised methods, as shown in Table \ref{supervised}. Our method demonstrates comparable performance to MCT-Net and even outperforms the supervised methods on the Pavia Center dataset. }

\begin{table}[t!]
   \centering
    \caption{ {Performance of different supervised methods on two datasets, and the downsampling ratio is 4 with 30 dB noise. The best results are \textbf{bold-faced}.}}
    \label{supervised}
    \setlength{\tabcolsep=0.01cm}{
    {
    \begin{tabular}{cccccc}
      \toprule
      &  Methods & MPSNR  & MSSIM   &SAM &  ERGAS  \\   \midrule
   \multirow{5}*{Pavia University}   
      & ConSSFCNN & 38.75 &0.97&  3.73    &   2.24     \\
       &  ResTFNet  & 40.70   &  0.97   &   3.00  &    1.94  \\
         & SSR-Net& 40.41    & 0.97  & 3.07     & 1.93     \\
         & MCT-Net & 42.10 & 0.98 & 2.68 &  1.62 \\
        & CLoRF   & \textbf{42.23} &  \textbf{0.99}& \textbf{2.05}& \textbf{1.31}\\
              \midrule
    \multirow{5}*{Pavia Center}   
       &   ConSSFCNN &  39.08 & 0.98 & 4.86 & 2.31  \\
       &   ResTFNet &  40.59  & 0.98 &   4.32   & 2.15 \\
         & SSR-NET & 40.19  & 0.98    & 4.07 & 2.14 \\
         & MCT-Net & 41.82 &  0.98  & 3.89 &1.86 \\
        &  CLoRF &  \textbf{43.47} & \textbf{0.99} & \textbf{3.64} & \textbf{1.20}  \\
          \bottomrule
    \end{tabular}
    }
    }
\end{table}

{5) Performance under degeneration with estimated PSF and SRF: To test the impact of degradation operators in the spatial and spectral domains on HR-HSI fusion results, we assume that the degradation operators are unknown in the simulated dataset, and we estimate these operators using the method suggested in \cite{simoes2014convex}. In Table \ref{blind}, we present the results of both semi-blind and fully-blind experiments. Note that the CNMF and HySure methods are fully blind in the experiment. They are not included in the semi-blind experiments for comparison. As shown in Table \ref{blind}, blind fusion is more challenging than semi-blind fusion, all methods perform similarly underestimated degradations in the spatial domain. However, in blind fusion, the performance of all methods decreases to some extent. Nevertheless, our method still demonstrates greater robustness in comparison to other approaches.}
\begin{table}[t!]
   \centering
    \caption{ {Performance under generation produced by semi-blind (estimate PSF) and blind (estimate both PSF and SRF) with  30 dB noise and the downsampling ratio is 4 for Pavia University. The best results are \textbf{bold-faced}.} }
    \label{blind}
    \footnotesize
     \tabcolsep=0.15cm
    {
    \begin{tabular}{cccccc }
      \toprule
      &  Methods & MPSNR  & MSSIM   &SAM &  ERGAS  \\   \midrule
   \multirow{4}*{Semi-blind}   
      & HyCoNet &  40.10& 0.97& 3.10& 1.41 \\
       & CNN-FUS  &  40.04 & 0.97& 2.55 & 1.64 \\
         & MSE-SURE   & 41.11 & 0.98  & \textbf{2.20} &1.49 \\
         & E2E-FUS   & 40.77 & \textbf{0.98}& 2.34 &1.52 \\
        & CLoRF   & \textbf{41.40} &  \textbf{0.98}& 2.32& \textbf{1.40}\\
              \midrule
    \multirow{7}*{Blind}   
          &  CNMF  &  32.70  & 0.92   & 3.55   & 3.63  \\
       &  HySure    &  32.86  &   0.93   & 5.81  & 3.71  \\
       &  HyCoNet    &  38.06 & 0.95  & 3.97  & \textbf{1.83} \\
       &  CNN-FUS    & 38.32 &   0.96 & 3.02 & 2.12 \\
         & MSE-SURE &   36.14 & 0.96  &  4.36& 2.90 \\
         & E2E-FUS     & 31.33  & 0.84 & 7.53 & 5.34\\
        &  CLoRF  &  \textbf{38.98} & \textbf{0.97}   & \textbf{3.00} & 1.88 \\
          \bottomrule
    \end{tabular}}
\end{table}

\subsection{Arbitrary Resolution}
\begin{figure}[htp]
\centering
\includegraphics[width=0.99\linewidth]{ 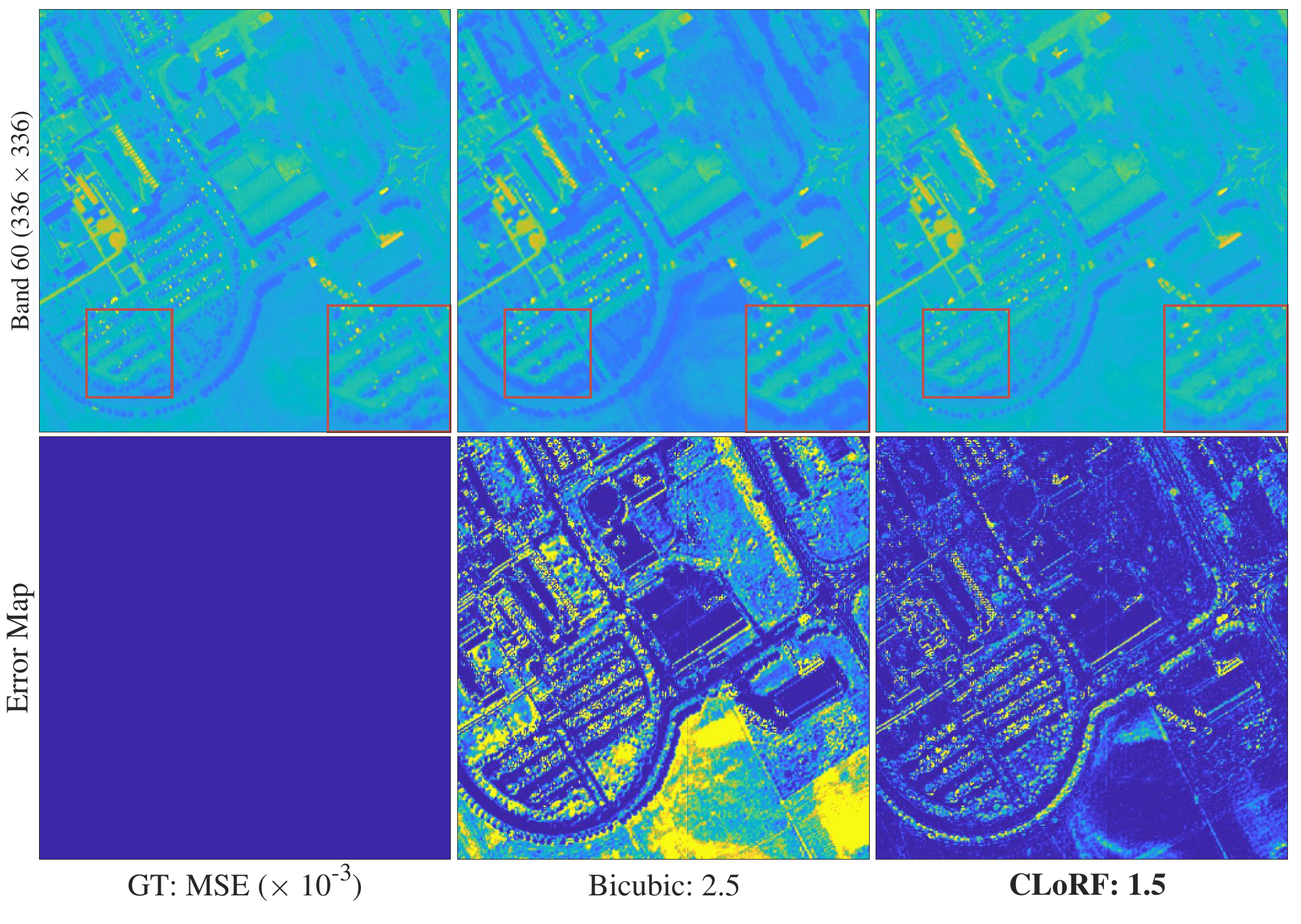}
  \caption{An example of
CLoRF for super-resolution on the Pavia Unversity with a spatial resolution of $336\times336$ (60th band) and its corresponding error map while simultaneously achieving arbitrary resolutions in both spatial and spectral domains. }
  \label{spatial_res}
\end{figure}

\begin{figure*}[htp]
\centering
\includegraphics[width=1.02\textwidth]{ 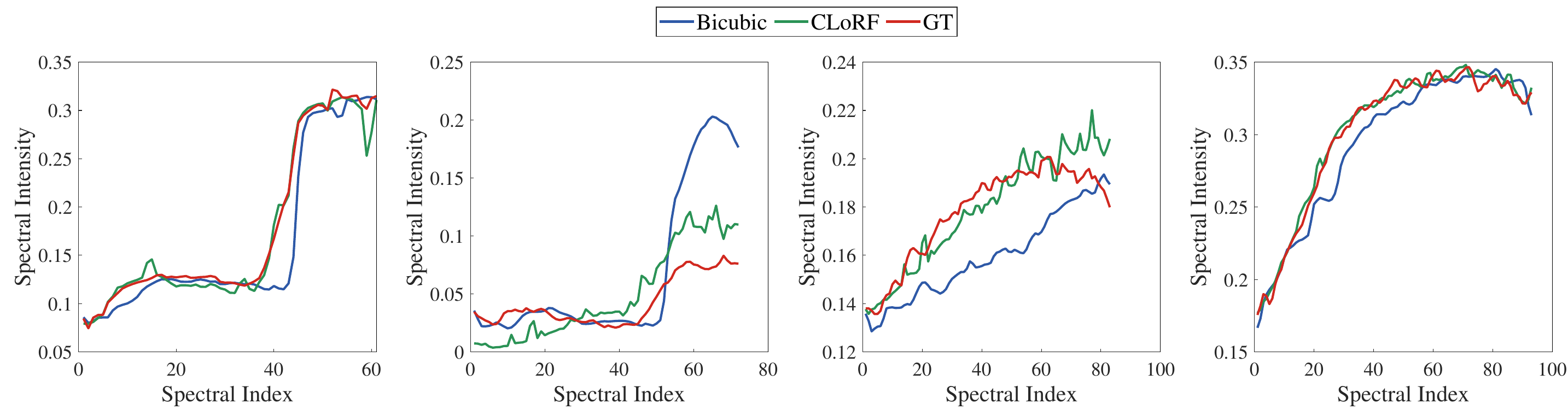}
  \caption{Visualize the spectral signatures of different pixels in different spectral resolutions while simultaneously obtaining arbitrary resolutions in both spatial and spectral domains.}
  \label{spectral_res}
\end{figure*}

\begin{figure*}[htp]
\centering
\includegraphics[width=0.99\linewidth]{ 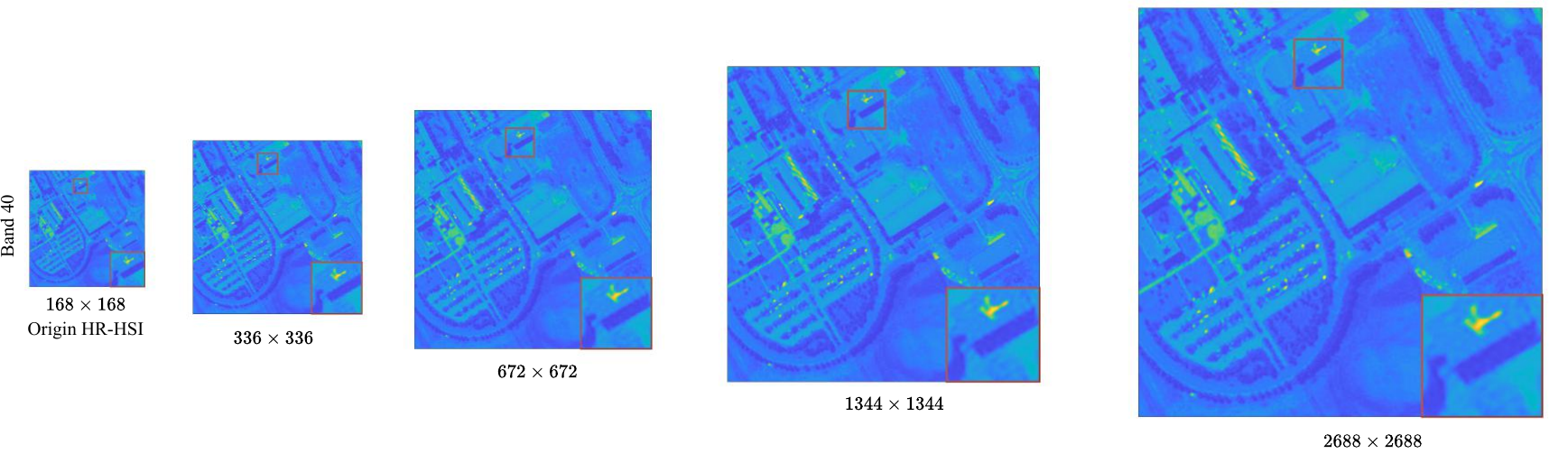}
  \caption{{The original resolution of HR-HSI is $(168,168,50)$, the spatial super-resolution factor in $\{2, 4, 8, 16\}$. Visualize the 40th band at different resolutions for Pavia University. }}
  \label{ari_res}
\end{figure*}

Here, we evaluate the performance of CLoRF to fuse HSIs at arbitrary spatial and spectral resolutions. As shown in Fig.~\ref{fig:aribrtary},  our model is well-trained based in an unsupervised manner, and we can input any scale spatial and spectral position coordinates during the inference stage, obtaining arbitrary resolutions in both spatial and spectral domains. 
We use PSF and downsampling ratio of the same size as in Sec.~\ref{experiment_detail}, but different sizes of SRF (sampled from SRF in Sec.~\ref{experiment_detail}) to synthesize the reduced-resolution LR-HSI ($42\times 42 \times 50$) and HR-MSI ($168\times 168 \times 4$) with Pavia University for training spatial-INR and spectral-INR. Then, we predict HR-HSIs of arbitrary resolutions using different scales of spatial and spectral coordinates. For a fair comparison, our method is directly compared with bicubic interpolation from the original GT. As shown in Table~\ref{arb_res}, we can fix spectral resolution to obtain arbitrary spatial resolution, fix spatial resolution to obtain arbitrary spectral resolution, or simultaneously achieve resolutions in both spatial and spectral domains. CLoRF performs nearly as well as bicubic interpolation in spatial resolution when spectral resolution is fixed, yet it notably surpasses bicubic when spatial resolution is fixed while achieving spectral resolution. Moreover, when simultaneously enhancing resolutions in both spatial and spectral domains, CLoRF consistently outperforms bicubic interpolation.
From Fig.~\ref{spatial_res}-\ref{spectral_res}, it can be seen that when simultaneously upsampling in both spatial and spectral domains, CLoRF obtains finer details compared to the bicubic interpolation. 
{Furthermore, we train a smaller size of data (LR-HSI ($42\times 42 \times 50$) and HR-MSI ($168\times 168 \times 4$) with Pavia University for training spatial-INR and spectral-INR) and infer the desired HR-HSI with spatial super-resolution factor in $\{2, 4, 8, 16\}$ and a fixed spectral upsampling resolution 93. 
    Figure \ref{ari_res} demonstrates that the trained CLoRF model successfully super-resolves the original HR-HSI $(168\times 168 \times 50)$ to arbitrary resolutions with a large range of flexibility. However, at a super-resolution factor of 16, the image exhibits slight fluctuations within some regions, which shows the inherent limitations.}
\begin{table*}[htp]
    \centering
      \caption{Experimental results in spatial (fixed spectral), spectral (fixed spatial), and (spatial, spectral) with arbitrary resolution, metric: MPSNR/MSSIM.
The original dimension of HR-HSI is (168, 168, 50). The best results are \textbf{bold-faced}. }
    \label{arb_res}
\footnotesize
    \begin{tabular}{cccccc}
    \toprule
    \multicolumn{5}{c}{Arbitrary resolution in the spectral domain  }\\  \midrule 
Dimension& (168,168,61) & (168,168,72) & (168,168, 83) & (168,168,93)  \\ 
 Bicubic&37.64 /0.97 & 36.60/0.95 & 38.76/0.96  & 37.23/0.95      \\
CLoRF& \textbf{42.42/0.99}& \textbf{42.85/0.98} & \textbf{42.64/0.98}& \textbf{42.56/0.98 }\\
    \midrule \midrule 
      \multicolumn{5}{c}{Arbitrary resolution in the spatial domain  }\\ \midrule 
Dimension & (210,210,50) & (252,252,50) & (294,294,50) &(336,336,50)  \\ 
Bicubic & \textbf{30.07/0.90} & \textbf{29.68/0.89} & \textbf{30.24/0.89}  & \textbf{29.29/0.88}\\  
 CLoRF& 29.86/0.88 & 29.52/0.87 & 30.04/0.86 & 29.22/0.86\\\midrule \midrule  
      \multicolumn{5}{c}{Arbitrary resolution in the spectral and spatial domain  }\\ \midrule 
 Dimension&  (210,210,61) & (252,252,72) & (294,294,83) &(336,336,93)  \\ 
  Bicubic   & 28.87/0.87 & 28.00/0.85 & 28.56/0.85       &27.48/0.83\\
   CLoRF& \textbf{29.53/0.88} & \textbf{29.21/0.87} & \textbf{29.60/0.86} & \textbf{28.61/0.84}\\  \bottomrule
  \end{tabular}
\end{table*}

\subsection{Ablation Study}
We conduct a series of ablation studies to verify the effectiveness of CLoRF.

1) Loss function: As shown in Table \ref{loss_fun},  incorporating TV loss into the loss function can lead to better recovery quality.  On each dataset, the TV loss is capable of enhancing PSNR by 1-2 dB.
\begin{table}[htbp]
    \centering
      \caption{Ablation study of the TV loss. The best results are \textbf{bold-faced}. }
    \label{loss_fun}
    \begin{tabular}{cccc}
    \toprule
     & Metric &   w/o  &    w/ \\
     \midrule
      \multirow{2}*{Pavia University}   & MPSNR &  41.48        & \textbf{42.23}\\
                                 & MSSIM &     0.98    &  \textbf{0.99}       \\
    \midrule 
     \multirow{2}*{University of Houston}   & MPSNR &    39.04     &        \textbf{40.55}     \\
                                 & MSSIM &   0.98      &  \textbf{0.99}    \\
    \bottomrule
  \end{tabular}
\end{table}

\begin{table}[htbp]
    \centering
     \caption{Ablation study of the different activation functions. (MPSNR / MSSIM). The best results are \textbf{bold-faced}.}
    \label{ablation_act}
    \begin{tabular}{ccc}
    \toprule
     Act.Func & Pavia University &  Houston \\ \midrule
     ReLU  &      25.92/0.65       &    24.58/0.54               \\
      ReLU+PE  &  36.26/0.93              &  37.56/0.98 \\
    Gauss &         35.96/0.90       &   36.10/0.96 \\
      Spder &       41.63/0.98     & 39.16/0.98\\
        Sine  &     \textbf{42.23/0.99}     &  \textbf{40.55/0.99} \\
    \bottomrule
  \end{tabular}
\end{table}

2) Different activation functions:
Table \ref{ablation_act} displays the results obtained using different activation functions, such as ReLU, ReLU+Position Encoding (PE), Gauss~\cite{ramasinghe2022beyond}, Spder~\cite{shah2023spder}, and Sine~\cite{sitzmann2020implicit}. Notably, the Sine activation function demonstrates outstanding performance across all metrics. Therefore, we have made it the default activation function for our model.

\begin{figure*}[thp]
  \centering
  \begin{subfigure}[b]{0.24\textwidth}
    \includegraphics[width=\textwidth]{ 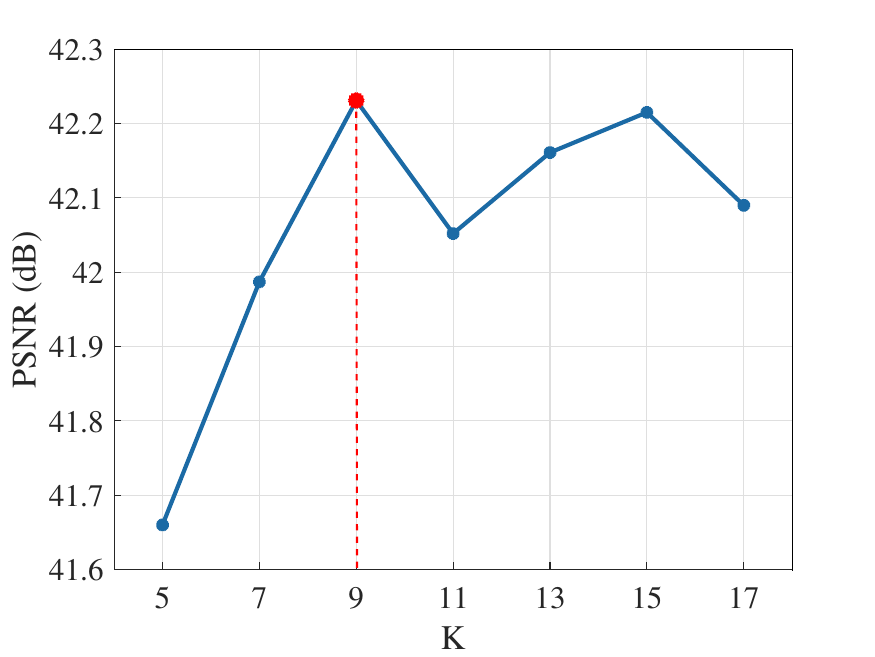}
  \end{subfigure}
  \begin{subfigure}[b]{0.24\textwidth}
    \includegraphics[width=\textwidth]{ 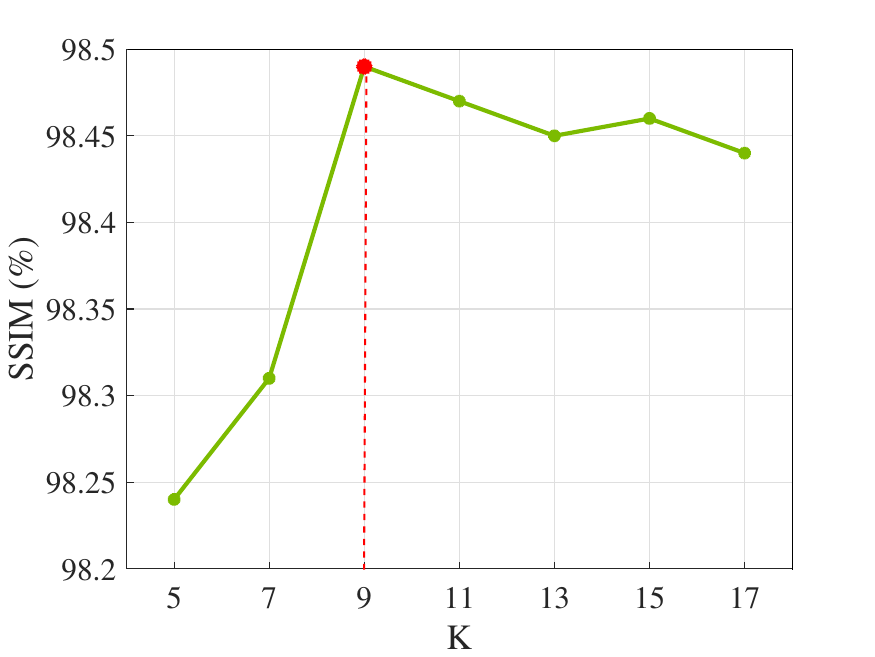}
  \end{subfigure}
  \begin{subfigure}[b]{0.24\textwidth}
    \includegraphics[width=\textwidth]{ 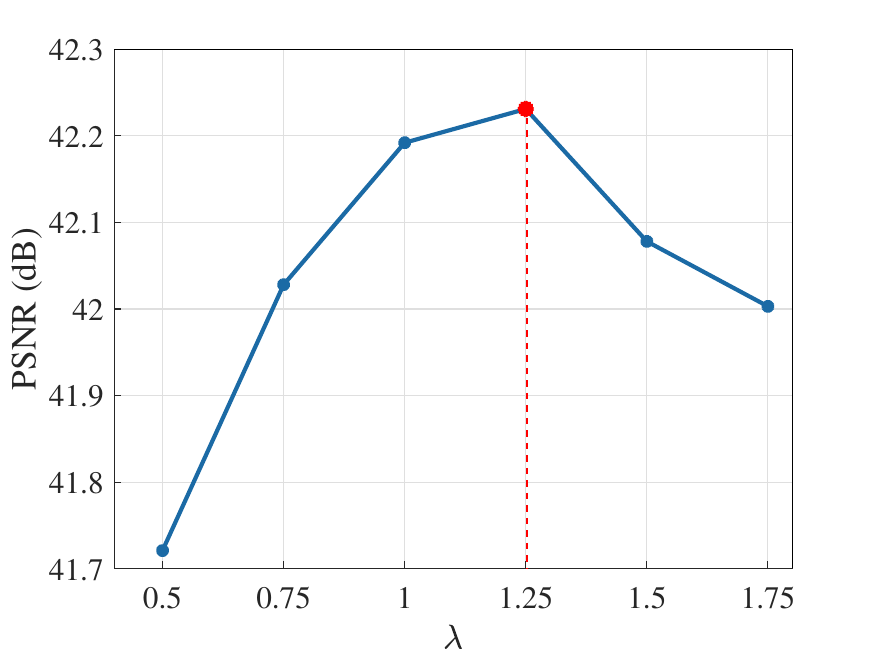}
  \end{subfigure}
  \begin{subfigure}[b]{0.24\textwidth}
    \includegraphics[width=\textwidth]{ 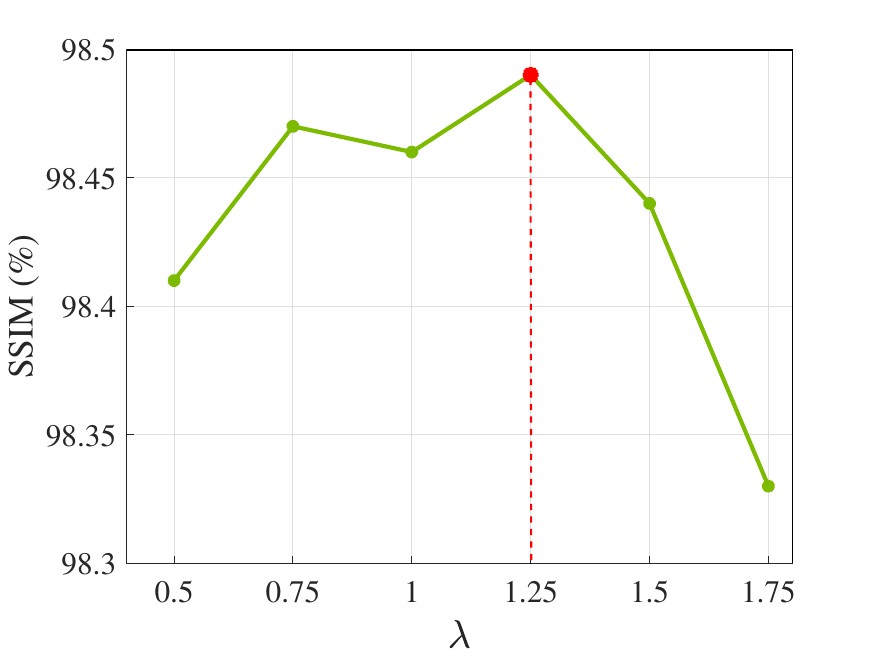}
  \end{subfigure}
  
  \begin{subfigure}[b]{0.24\textwidth}
    \includegraphics[width=\textwidth]{ 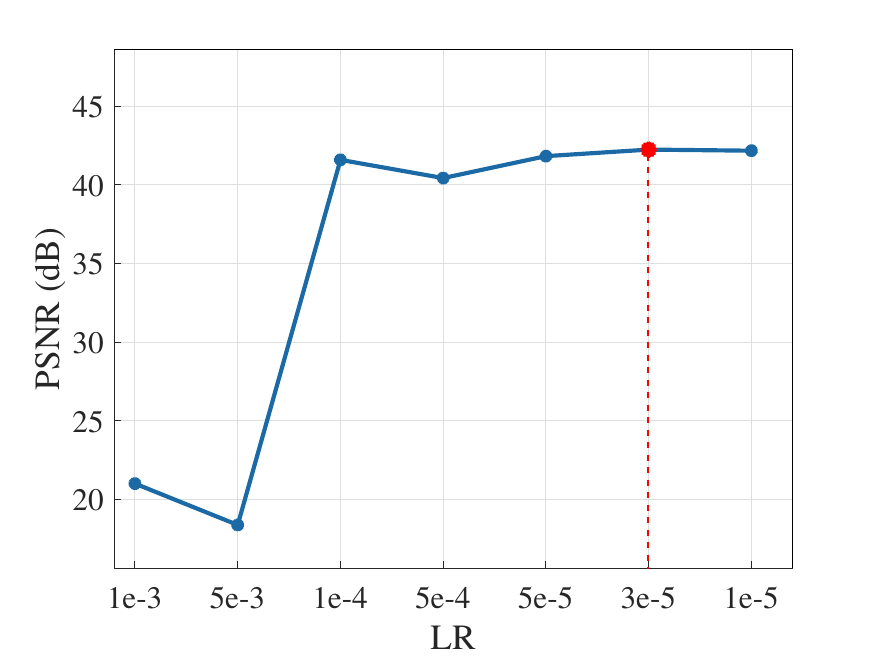}
  \end{subfigure}
  \begin{subfigure}[b]{0.24\textwidth}
    \includegraphics[width=\textwidth]{ 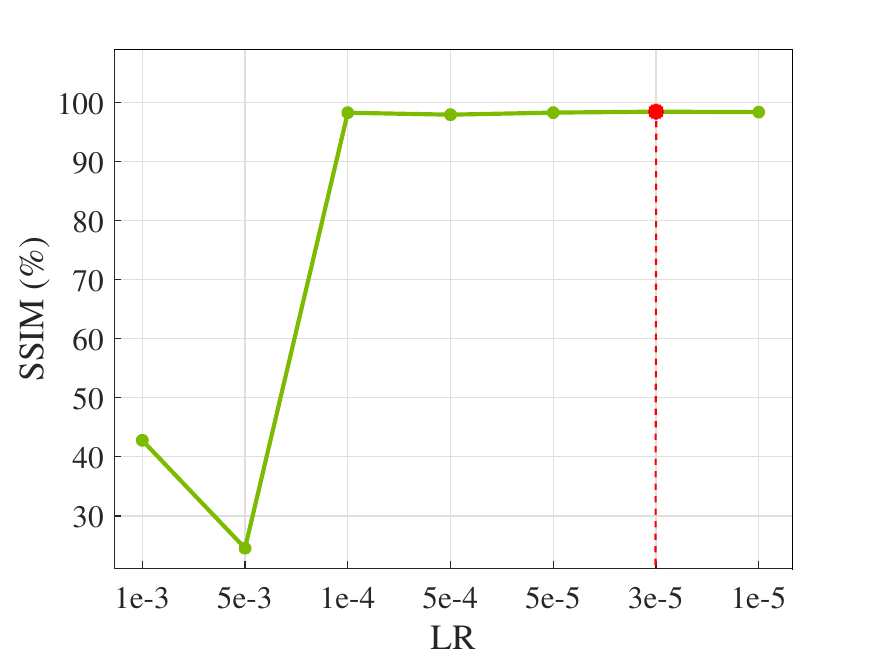}
  \end{subfigure}
  \begin{subfigure}[b]{0.24\textwidth}
    \includegraphics[width=\textwidth]{ 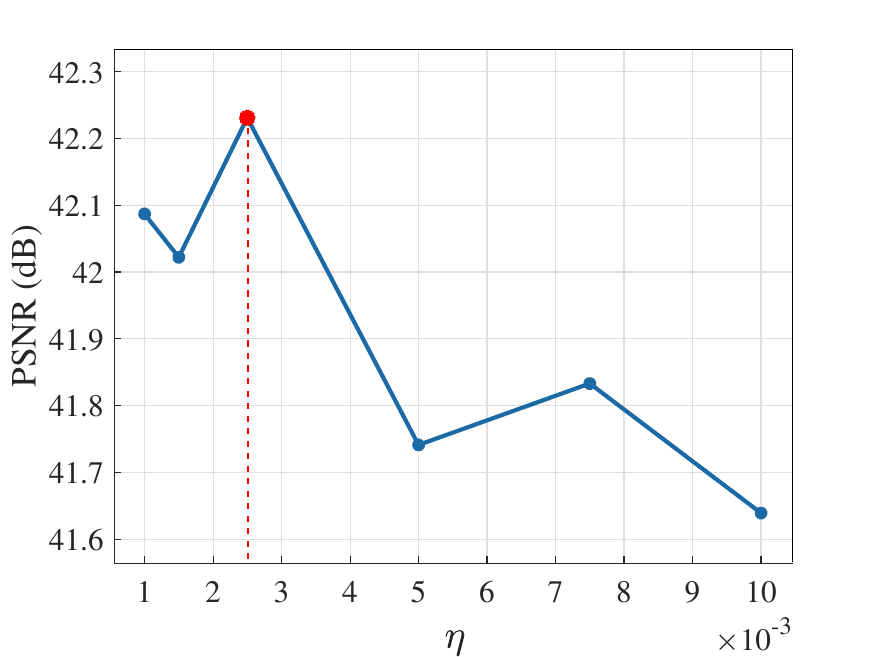}
  \end{subfigure}
  \begin{subfigure}[b]{0.24\textwidth}
    \includegraphics[width=\textwidth]{ 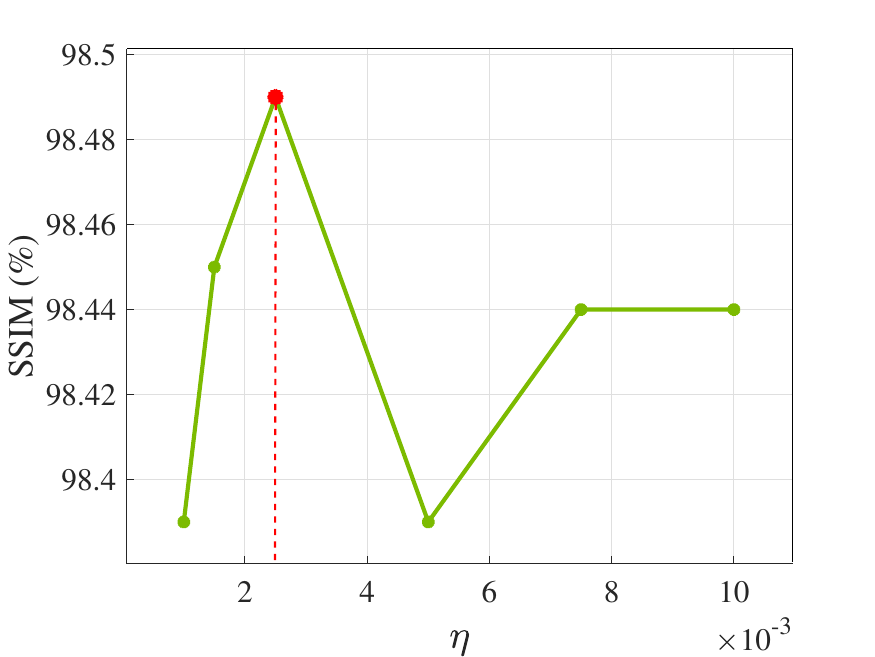}
  \end{subfigure}

  \begin{subfigure}[b]{0.24\textwidth}
    \includegraphics[width=\textwidth]{ 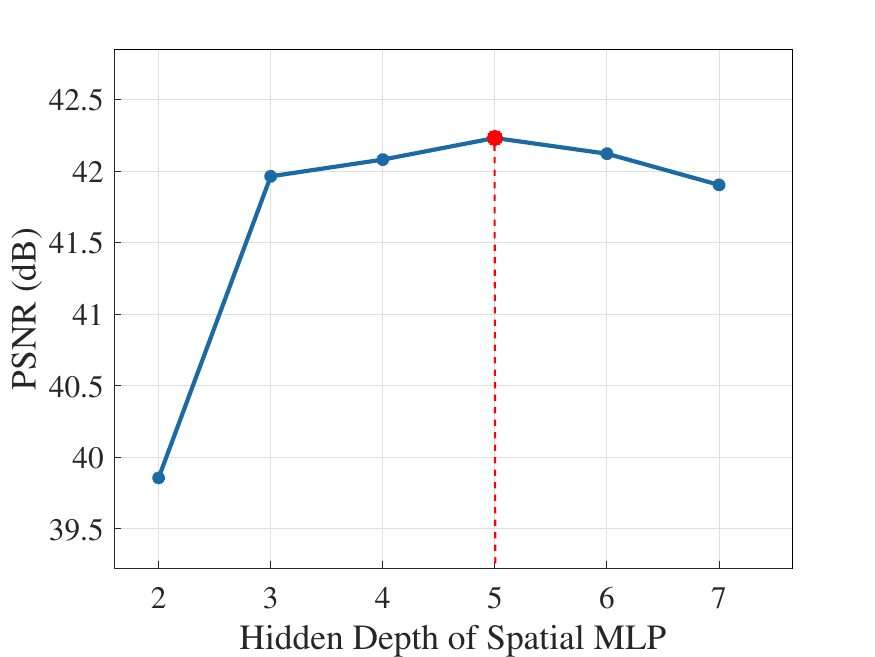}
  \end{subfigure}
  \begin{subfigure}[b]{0.24\textwidth}
    \includegraphics[width=\textwidth]{ 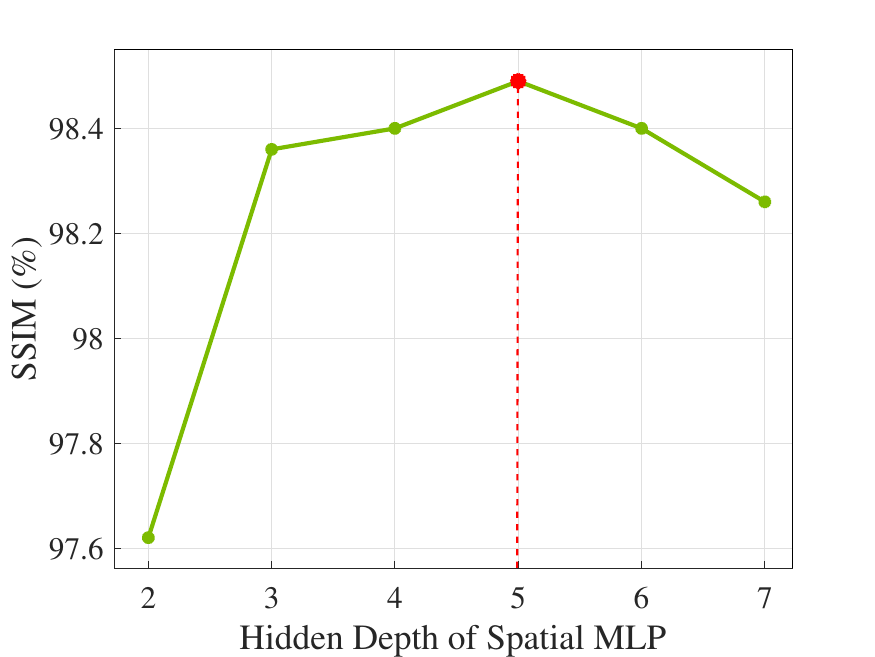}
  \end{subfigure}
  \begin{subfigure}[b]{0.24\textwidth}
    \includegraphics[width=\textwidth]{ 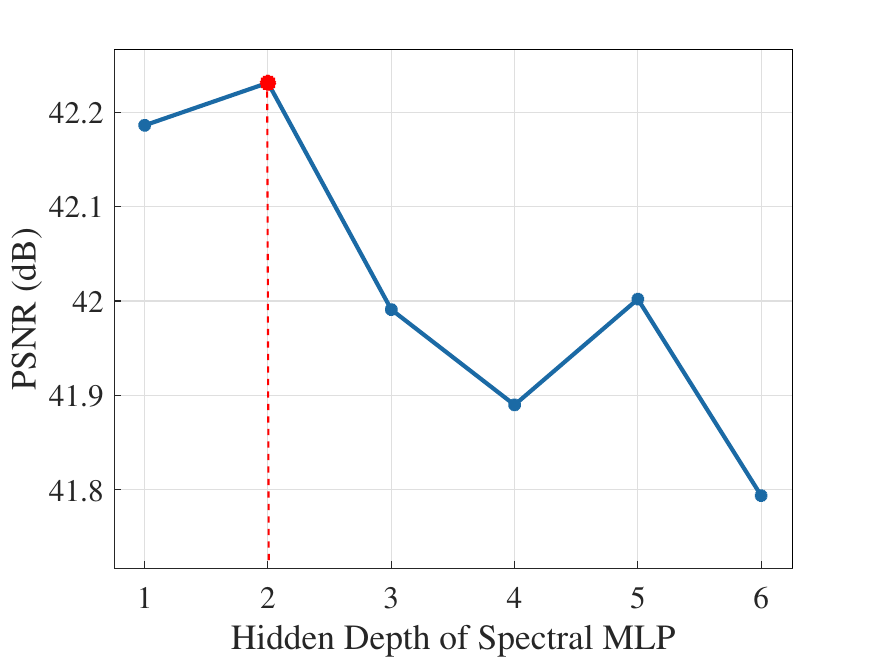}
  \end{subfigure}
  \begin{subfigure}[b]{0.24\textwidth}
    \includegraphics[width=\textwidth]{ 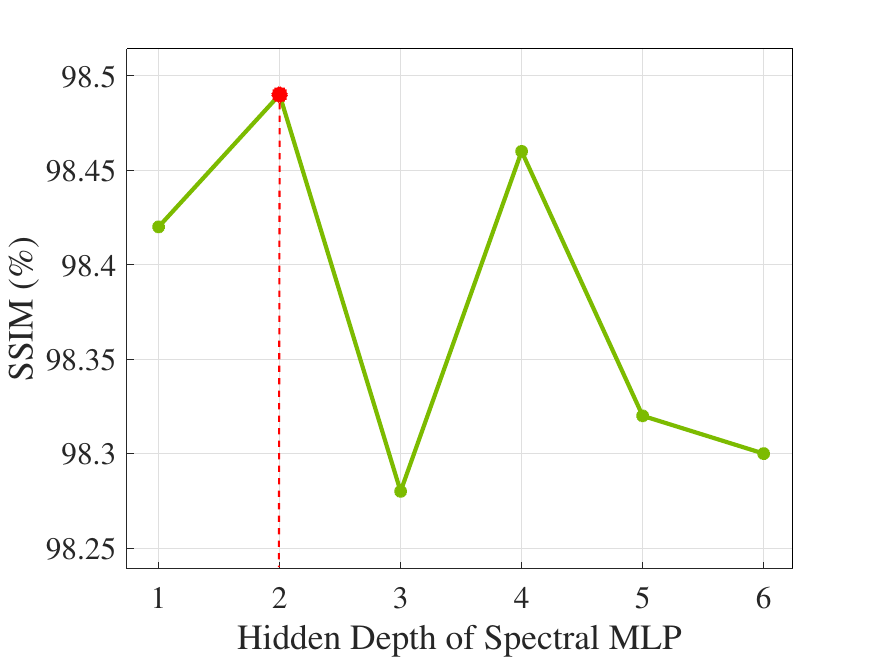}
  \end{subfigure}
  \caption{Impact of hyperparameters on MPSNR and MSSIM. 
The red dot represents the best result obtained by traversing all values.}
  \label{ablation_para_combined}
\end{figure*}

3) Impact of hyper-parameters: We examine six hyperparameters:  $K$,  $\lambda$, $\eta$, the hidden depth of two MLPs, and the learning rate (LR). Each hyperparameter is explored within a specified range while the others are held constant. Specifically, for Pavia University, we explore $K$ within the range of 5 to 17. For $\lambda$, we consider values from the set $\{0.5,0.75,1,1.25,1.5, 1.75\}$, For $\eta$, the search space consists of $\{ 10^{-3}, 2.5\times 10^{-3}, 5\times 10^{-3},7.5\times 10^{-3}, 10^{-2}\}$. Regarding the hidden depth of the two MLPs, we initially set the hidden depth of the spectral MLP at 2 and explored the depth of the spatial MLP from 1 to 7. Similarly, we fix the hidden depth of the spatial MLP at 5 and then search for the hidden depth of the spectral MLP from 1 to 6. For LR, we consider values from the set $\{10^3,5\times 10^3,10^4, 5\times 10^4, 10^5, 3\times 10^5,5\times 10^5\}$. The results are displayed in Fig.~\ref{ablation_para_combined}.


\subsection{Evaluation on Real Dataset}
{To further illustrate the effectiveness of the proposed method, we conduct a real-world fusion experiment. 
It is important to note that the ground-truth spatial and spectral degradation operators are not available for real data. Consequently, we estimate these operators using the approach suggested in \cite{simoes2014convex}. For parameter selection in real data, we first apply the classical noise estimation algorithm proposed in \cite{bioucas2008hyperspectral} to obtain a preliminary estimate of the noise intensity of the LR-HSI, which is SNR=32.75 dB.  Since the estimated noise intensity is close to SNR = 30 dB,  In our method, we set $K$ and $\eta$ to be consistent with Pavia University, and  $\lambda=1.8$. The experimental parameters for the other methods remain consistent with those of the Pavia University dataset. }

\begin{figure*}[t!]
\begin{center}
\includegraphics[width=1\textwidth]{ 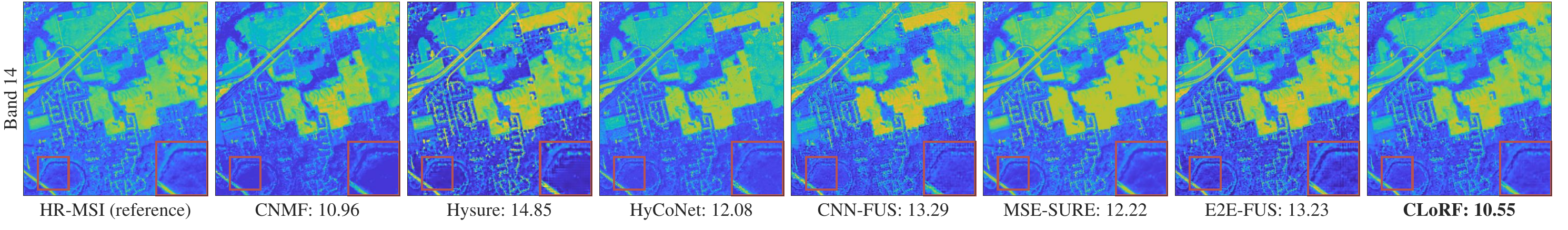} 
\end{center}
\caption{ {Visual the real data (14th band) of the estimated HR-HSI, evaluated with a no-reference hyperspectral imaging quality score for various methods applied to the real HypSen dataset in blind fusion tasks.}}
\label{real}
\end{figure*}

{Since there is no ground-truth for the HR-HSI, we visualize the estimated HR-HSIs of all methods in Fig. \ref{real}  together with the image quality score measured by a non-reference image quality metric~\cite{yang2017no}. We find that the proposed CLoRF recovers more details and obtains the best image quality.}

\subsection{PAN-HSI Fusion}
In this section, we extend our proposed fusion method to the PAN-HSI fusion. PAN image, compared to MSI, has fewer bands, making fusion more challenging. Below are the results and details of some experiments we conducted.

1) Dataset: The dataset acquired for the PRISMA contest1~\cite{vivone2022panchromatic}, namely RR1. The image is of size $900\times 900 \times 59 $. The synthetic datasets, plus a simulated PSF and SRF, were used to generate two observed images: LR-HSI and PAN. PSF is set to be the same as the~\ref{experiment_detail}. The wavelength~\cite{vivone2022panchromatic} was used to simulate the SRF to generate the PAN image.  The i.i.d Gaussian noise is added to the LR-HSI (30 dB) and PAN (30 dB) image.

2) Compared Methods:  We primarily compare CLoRF with two categories of baselines: model-based methods include GS~\cite{laben2000process}, GSA~\cite{aiazzi2007improving}, AWLP~\cite{vivone2014critical}, MTF-GLP~\cite{aiazzi2006mtf}, and HySure~\cite{simoes2014convex}, while the unsupervised deep learning method is R-PNN~\cite{guarino2023band}. All training parameters for CLoRF remain the same for Pavia University.

\begin{figure*}[thp]
\begin{center}
\includegraphics[width=1\textwidth]{ 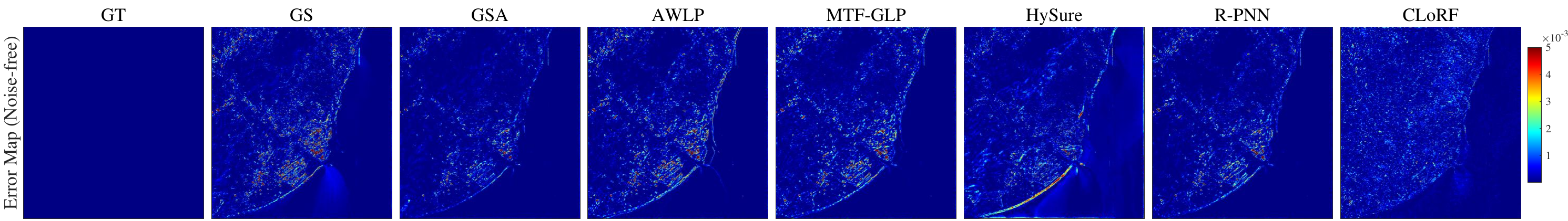} 
\includegraphics[width=1\textwidth]{ 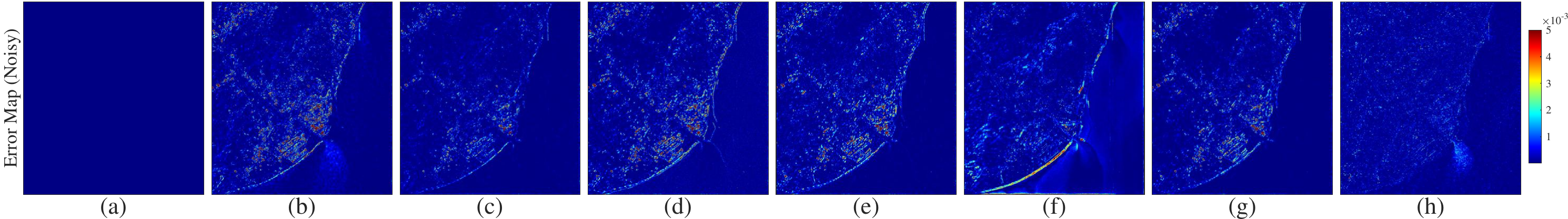} 
\end{center}
\caption{The first and second rows show the error images (Noise-free and Noisy) between the estimated HR-HSI and GT.}
\label{PAN_compare}
\end{figure*}

As shown in Table~\ref{PAN_RR1}, CLoRF outperforms the other methods in terms of MPSNR and ERGAS. However, due to spectral distortion in the PAN image, CLoRF fails to learn a continuous representation in the spectral domain effectively.
For clarity, we depict error maps for two scenarios: noise-free and noisy, as shown in Fig.~\ref{PAN_compare}. The results indicate that CLoRF outperforms other methods in spatial domains.
\begin{table*}[thp]
    \centering
        \caption{Quantitative performance comparison with different algorithms on the RR1 dataset. The best results are \textbf{bold-faced}, and runner-ups are \underline{underlined}. (MPSNR \textuparrow, MSSIM \textuparrow, SAM \textdownarrow, ERGAS \textdownarrow). }
    \label{PAN_RR1}
       \footnotesize
    \begin{tabular}{ccccccccc}
      \toprule
      &  Metric & GS &  GSA &AWLP  & MTF-GLP  &HySure   &R-PNN  &CLoRF  \\ \midrule
   \multirow{4}*{Noise-free}   
       & MPSNR & 27.40  &\underline{32.25}&29.64&30.13&29.56&30.05&\textbf{32.87}\\
         &  MSSIM &0.84&\textbf{0.92}&0.90&0.90 & 0.88 &0.90 &\underline{0.91} \\
         &   SAM &9.70 & \underline{4.46} & 4.90 &\textbf{4.41}& 6.49 &4.78& 5.81 \\
        &  ERGAS&  5.27 & \underline{3.06} &3.92 &3.75& 4.27 &4.28 &\textbf{2.95}\\\midrule
 &  Metric & GS &  GSA &AWLP  & MTF-GLP  &HySure   &R-PNN  &CLoRF  \\ \midrule
   \multirow{4}*{Noisy}   
       & MPSNR & 27.25& \underline{31.90}&29.23&29.74&29.44& 29.33 &\textbf{32.61}\\
         &  MSSIM &0.82&\textbf{0.90}&0.87&0.88 & 0.88 &0.88 &\underline{0.89} \\
         &   SAM &10.05& \textbf{5.51}&5.96& \underline{5.76}& 7.07 &6.05& 6.50 \\
        &  ERGAS&  5.35&\underline{3.16} &4.08 &3.90&4.12 &4.60 &\textbf{2.99}\\
               \bottomrule
   \end{tabular}
\end{table*}
\section{Conclusion}\label{sec:conclusion}
In this work, we introduced an innovative continuous low-rank factorization representation for HSI-MSI fusion, which incorporates two INRs into the decomposition to capture spatial and spectral information, respectively. Theoretical analysis reveals that this continuous function representation adeptly portrays the low-rank and smoothness priors of HSIs. Extensive numerical experiments conducted for HSI-MSI fusion confirm its effectiveness and wide applicability. Nonetheless, our method still faces certain limitations. Given its unsupervised manner, CLoRF demands an extensive number of training epochs.  Another limitation is that we solely utilize SIREN, which is widely used in INR. There's a need to explore diverse continuous representations for both spatial and spectral domains.
The adoption of continuous low-rank factorization representation for processing and analyzing HSIs shows potential for future applications across various tasks, e.g., HSI unmixing and single RGB-HSI super-resolution.  Our future efforts will be dedicated to expanding the versatility of CLoRF to address diverse HSI tasks.



\bmhead{Acknowledgements}

This work was partly supported by 
the Natural Science Foundation
of China (Nos. 12201286, 52303301), the  Shenzhen Science and Technology Program (20231115165836001), HKRGC Grant No.CityU11301120, CityU Grant No. 9229120, HKRGC GRF 17201020 and 17300021, HKRGC C7004-21GF, and Joint NSFC and RGC N-HKU769/21, National Key R\&D Program of China (2023YFA1011400), and the Shenzhen Fundamental Research Program (JCYJ20220818100602005).

\bmhead{Data Availability}
Data will be made available on reasonable request.

\begin{appendices}
\section{Proof of Proposition 1} \label{sec:proof_prop1}
     Since $\h X\in S[f]$, we directly get MF-rank$[f]\geq \mathrm{rank}({\h X})$.
    Now, we aim to prove the other side: MF-rank$[f]\leq \mathrm{rank}({\h X})$. 

    Let $\mathbf{M}$ be any matrix within the set $S[f]$. Each column vector of $\mathbf{M}$ is denoted by $\mathbf{M}_{(:,p)}$ for $p \in \{1, 2, \ldots, n_2\}$. According to the definition of $S[f]$, there exists an index $l_p \in \{1, 2, \ldots, n_1\}$ dependent on $p$ such that $\mathbf{M}_{(:,p)}$ is a permutation of the elements in $\mathbf{X}_{(:, l_p)}$, allowing for repeated sampling.
In other words, for each $\mathbf{M}_{(:,p)}$, there exists a permutation matrix $\mathbf{P} \in \{0,1\}^{n_1 \times n_1}$ and a corresponding column of $\mathbf{X}$ depending on $p$ (specifically $\mathbf{X}_{(:, l_p)}$), such that $\mathbf{M}_{(:,p)} = \mathbf{P} \mathbf{X}_{(:, l_p)}$.
Additionally, the permutation matrix $\mathbf{P}$ is consistent across all columns $\mathbf{M}_{(:,p)}$ for $p = 1, 2, \ldots, n_2$, i.e., $\mathbf{M}_{(:,p)} = \mathbf{P} \mathbf{X}_{(:, l_p)}$ for each $p$.

Define $\tilde{\mathbf{X}} := [\mathbf{X}_{(:, l_1)}, \mathbf{X}_{(:, l_2)}, \ldots, \mathbf{X}_{(:, l_{n_2})}] \in \mathbb{R}^{m_1 \times n_2}$, we have that the rank of $\tilde{\mathbf{X}}$ is less than or equal to the rank of $\mathbf{X}$: $\mathrm{rank}(\tilde{\mathbf{X}}) \leq \mathrm{rank}(\mathbf{X})$.
Finally, since $\mathbf{M} = \mathbf{P} \tilde{\mathbf{X}}$, it follows that $\mathrm{rank}(\mathbf{M}) \leq \mathrm{rank}(\mathbf{X})$, which leads to MF-rank$[f]\leq \mathrm{rank}({\h X})$. 

\section{Proof of Theorem 3} \label{sec:proof_thm3}
First, we establish a linear representation for each factor function $f(\mathbf{s}, b)$ with fixed inputs $\mathbf{s}$ and $b$, employing a set of basis functions. Then, we focus on presenting the continuous low-rank factorization and demonstrating that this factorization preserves the matrix factorization rank (MF-rank). 

Suppose that MF-rank$[f]=K$ with $K<\infty$,
thus there exist a matrix $\h M \in \mathbb R^{n_1\times K}$ with rank$(M)= K$. Denote:  $\mathcal{S} = \{\h  s_i \ | \  \h M_{i,j}= f(\h s_{i},b_j), i= 1, \dots, n_1 \}, $ and $\mathcal{T} = \{ b_j \ | \ \h M_{i,j}= f(\h s_{i},b_j), j= 1, \dots, K\}. $ It is easy to see that $\{\h M_{(:,i)}\}_{i=1}^K$ are the column basis of $S[f]\bigcup \mathbb R^{n_1}$.  

Given any matrix $\h U \in \mathbb{R}^{n_1\times n_2} ( n_2 \geq K)$ as $\h U_{(i,j)}= f(\h s_{i}, b_j)$ with $\h s_{i}\in \mathcal{S}, b_j \in \mathcal{Z}_f$, we have $\h U \in S[f]$ and $\mathrm{rank}(\h U)\leq K$. Furthermore, the column vector in $\h U$ is a linear combination of the column basis $ \{\h M_{(:,i)}\}_{i=1}^K$: 
\begin{equation}
\label{eq:u_b_j}
    \h U_{(:,j)}=\sum_{k=1}^K c_{k}^{(b_j)} \h M_{(:,k)}, \text{ for } j=1,2, \dots, n_2. 
\end{equation}
Here,  we utilize Eq.\eqref{eq:u_b_j} and rewrite  $f(\h s, b)$ by  $\h c^{(b)} = [c^{(b)}_1, c^{(b)}_2, \dots , c^{(b)}_K]$:
\begin{equation}
f(\h s,  b)=\sum_{k=1}^K   c_{k}^{(b)} \h M_{(i,k)}=\sum_{k=1}^K c_{k}^{(b)} f(\h s, b_k), 
\end{equation}
for any  $\h s\in \mathcal{S},  b \in \mathcal{Z}_f. $

Next, we will generalize this conclusion from $\h \h \h {s} \in \mathcal{S}$ to any $\h {s} \in \mathcal{A}_f$. 
Given $ \h {\Tilde{s}} \in \mathcal{A}_f / \mathcal{S}$  and we construct a  matrix: $\h T \in \mathbb{R}^{(n_1+1)\times n_2}$, where $\h T_{(i,j)}=f(\h s_{i},b_j)$ and $\h s_i \in \mathcal{S} $ for $i=1, 2, \dots, n_1$ and $\h s_{n_1+1} = \h {\Tilde{s}}$. Assume there exist $K$ column vectors $\{\h T_{(:,j_k)}\}_{k=1}^K$ such that
$\h T_{(1:n_1,j_k)}=\h M_{(:,k)}, $ for $k=1, 2, \dots, K$. 
Hence, we get that $\mathrm{rank}(\h T) =K$ and  for $ j=1,2, \dots, n_2, $   
$$
\h T_{(:,j)}=\sum_{k=1}^K d_{k}^{(b_j)} \h T_{(:,j_k)},
$$

$$
 \h T_{(1:n_1,j)}=\sum_{k=1}^K c_{k}^{(b_j)} \h M_{(:,k)}.
$$
Owing to the uniqueness of the coefficient vector, we get that $\h d^{(b_j)} = \h c^{(b_j)}$. Hence, we have
\begin{equation}
\label{linear6}
\h T_{(n_1+1,j)}=\sum_{k=1}^K  c_{k}^{(b_j)} \h T_{(n_1+1,k)},
\end{equation}
which leads 
$
f(\h {\Tilde{s}},b)=\sum_{k=1}^K  c_{k}^{(b)} f(\h {\Tilde{s}}, b_{k})
$
 for any $\h{\Tilde{s}}\in \mathcal{A}_f / \mathcal{S}$. This gives the linear representation form of the factor function $f(\h s, b)$ (with fixed $\h s$ and $b$) using some basis functions $f(\h s,  b_{k})$ with $b_{k}\in \mathcal{T}$.


We define the factor function $f_{\mathrm{spatial}}(\cdot): \mathcal{A}_f \rightarrow \mathbb{R}^K$ as 
$$
f_{\mathrm{spatial}}(\h {\Tilde{s}}): =[f(\h {\Tilde{s}}, b_{1}),f(\h {\Tilde{s}}, b_{2}),\dots, f(\h {\Tilde{s}},  b_{K})]^T.
$$
Also, define the matrix function $h(\cdot): \ \mathcal{N}^{(K)} \times \mathcal{Z}_f
\rightarrow \mathbb{R}$ as
$$
h(i,b):=c_{i}^{(b)},
$$
where $\mathcal{N}^{(K)}  = \{1,2,\dots, K\}. $ 
From the above analysis, we  see that for any $(\h s,b) \in \mathcal{D}_f =\mathcal{A}_f \times \mathcal{Z}_f$, it holds that
\begin{equation}
\label{eq:fst}
    f(\h s,b)=\sum_{k=1}^K h(k,b) (f_{\mathrm{spatial}}(\h s))_{(k)}.
\end{equation}
Denote 
$f_{\mathrm{spectral}}(\cdot): \mathcal{Z}_f \rightarrow \mathbb{R}^K$ as 
$$f_{\mathrm{spectral}}(b):=[h(1,b),h(2,b),\cdots, h(K,b)]^T \in \mathbb{R}^{K\times 1}, $$ 
then Eq.\eqref{eq:fst} is rewritten as 
\begin{equation}
\label{eq:fst_final}
    f(\h s,b) = f_{\mathrm{spatial}}(\h {s}) \cdot f^T_{\mathrm{spectral}}(b).
\end{equation}
\section{Proof of Theorem 4} \label{sec:proof_thm4}
For any $(\h s_1 , b_1), (\h s_2, b_2) \in \mathcal{D}_f$, we have 
\begin{equation}
\begin{aligned}
     &|f(\h s_1, b_1)-f(\h s_2, b_1)| \\
     = & |\Phi_{\alpha}(\h s_1) \cdot\Psi_{\theta}^T(b_1)-\Phi_{\alpha}(\h s_2) \cdot \Psi_{\theta}^T(b_1)|\\
     \leq& | (\Phi_{\alpha}(\h s_1)-\Phi_{\alpha}(\h s_2))\cdot \Psi_{\theta}^T(b_1)|\\
     \leq&  \| \Phi_{\alpha}(\h s_1)-\Phi_{\alpha}(\h s_2) \|_{1}\|\Psi_{\theta}^T(b_1)\|_{1}.
\end{aligned} 
\end{equation}
Note that $\sigma(\cdot)$ is Lipschitz continuous,  i.e., $|\sigma(x)-\sigma(y)| \leq \kappa |x-y|$ holds for any $x,y,$ and letting
$y=0$ derives $\sigma(x) \leq \kappa |x|$ since $\sigma(0)=0$. 
On the other hand, denote $\psi^{(1)}(b) =  \h W_{1}^1 t$ and $\psi^{(k)}(b)=  \h W_{k}^1 \sigma (\psi^{(k-1)}(b))$, and {$\|\h W_i^1\|_1, \|\h W_i^2\|_1$ are bounded by a positive constant $\eta$ for all $i$}. So we get:
\begin{equation}
\label{eq:psi}
\begin{aligned}
& \|\Psi_{\theta}(b)\|_{1} = \| \psi^{(d)}(b)\|_{1} \\ \leq &  \|\h W_{d}^1\|_1 \|\sigma (\psi^{(d-1)}(b)) \|_1\\
 \leq & \eta \kappa \|\psi^{(d-1)}(b)\|_{1}
\leq  \eta^{d}\kappa^{d-1} |b|.
\end{aligned}
\end{equation}
Meanwhile, we denote $\phi^{(1)}(\h s) =  \h W_{1}^2 \h s$ and $\phi^{(k)}(\h s)=  \h W_{k}^2 \sigma (\phi^{(k-1)}(\h s))$. Then 
it holds that
\begin{equation}
\label{eq:phi}
\begin{aligned}
   & \|\Phi_{\alpha}(\h s_1)-\Phi_{\alpha}(\h s_2)\|_{1}\\
  =&\| \phi^{(d)}(\h s_1)-\phi^{(k)}(\h s_2)\|_{1}\\
  =&\| \h W_d^2 (\sigma (\phi^{(d-1)}(\h s_1))-\sigma (\phi^{(d)}(\h s_2))\|_{1}\\
   \leq &\eta \kappa \| \phi^{(d-1)}(\h s_1)-\phi^{(d-1)}(\h s_2)\|_{1}\\
   \leq & \eta^{d}\kappa^{d-1} \|\h s_1-\h s_2\|_{1},
\end{aligned}
\end{equation}
{Let $\zeta=\max\{\|\h s_1\|_{1},|b_1|\}$}. Combining the inequalities Eq.\eqref{eq:psi} and Eq.\eqref{eq:phi}, we have

\begin{equation}
\begin{aligned}
    &| f(\h s_1,b_1)- f(\h s_2,b_1)| \\ 
    \leq  &\eta^{2d+1}\kappa^{2d-2} |b_1| \|\h s_1-\h s_2\|_{1}\\
   \leq &\eta^{2d+1}\kappa^{2d-2} \zeta \|\h s_1-\h s_2\|_{1}.
\end{aligned}
\end{equation}

Similarly, we prove that 
\begin{equation*}
    |f( \h s_2,b_1)- f(\h s_2,b_2)| \leq \eta^{2d+1}\kappa^{2d-2} \zeta|b_1-b_2|.
\end{equation*}
Combining the above two inequalities, we get
{\begin{equation*}
   \begin{split}
       |f(\h s_1,b_1)-f(\h s_2,b_2)|  
         \leq & |f(\h s_1, b_1)-f(\h s_2,  b_1)| 
        \\ & +|f(\h s_2,  b_1)-f(\h s_2, b_2)| \\ 
         \leq &  \delta\|\h s_1-\h s_2\|_{1} +\delta|b_1-b_2|, 
   \end{split}
\end{equation*}
where $\delta = \eta^{2d+1}\kappa^{2d-2} \zeta. $ }
\end{appendices}

\bibliographystyle{apacite}
\bibliography{sn-bibliography}

\end{document}